\documentclass{article} 
\usepackage{iclr2026_conference,times}

\usepackage{amsmath,amsfonts,bm}

\def\eqref#1{equation~\ref{#1}}

\def\1{\bm{1}}

\DeclareMathAlphabet{\mathsfit}{\encodingdefault}{\sfdefault}{m}{sl}
\SetMathAlphabet{\mathsfit}{bold}{\encodingdefault}{\sfdefault}{bx}{n}

\usepackage{microtype}
\usepackage{graphicx}
\usepackage{subcaption}
\usepackage{booktabs} 
\usepackage{algcompatible}
\usepackage{algpseudocode}
\usepackage{hyperref}
\usepackage{url}
\usepackage{amsmath}
\usepackage{amssymb}
\usepackage{mathtools}
\usepackage{amsthm}
\usepackage{makecell}
\usepackage{multirow}

\usepackage{tcolorbox}
\usepackage{xcolor}
\definecolor{examplebg}{RGB}{245,247,250}
\tcbuselibrary{breakable}

\usepackage{hyperref}
\usepackage{caption}
\usepackage{longtable}
\usepackage{enumitem}
\usepackage{tcolorbox}
\usepackage{listings}
\usepackage{tcolorbox}
\tcbuselibrary{skins,breakable}
\usepackage{booktabs}
\usepackage{longtable}
\usepackage{enumitem}
\usepackage{verbatim}
\tcbuselibrary{breakable}

\usepackage[capitalize,noabbrev]{cleveref}

\theoremstyle{plain}

\theoremstyle{definition}

\theoremstyle{remark}

\usepackage{xspace}

\usepackage[textsize=tiny]{todonotes}
\usepackage[table]{xcolor}
\usepackage{colortbl}

\title{Scaling Agentic Capabilities, Not Context: \\Efficient Reinforcement Finetuning for Large Toolspaces}

\author{%
 Karan Gupta, Pranav Vajreshwari, Yash Pandya, Raghav Magazine, Akshay Nambi, Ahmed Awadallah \\
  \textbf{Microsoft Research} \\
  \textit{Corresponding author:~akshayn@microsoft.com} \\
 }

\newcommand{\pname}{\texttt{ATLAS}\xspace}
\iclrfinalcopy 
\begin{document}

\maketitle

\begin{abstract}
Agentic systems operating over large tool ecosystems must plan and execute long-horizon workflows under weak or non-verifiable supervision. While frontier models mitigate these challenges through scale and large context budgets, small language models (SLMs) remain brittle: eager tool loading saturates context, execution errors compound over time, and sparse rewards limit learning.
We introduce \pname, a reinforcement finetuning framework that enables SLMs to operate effectively in large-scale toolspace environments by learning \emph{how} to acquire context and \emph{how} to execute actions. Our approach makes two key contributions. First, we treat context control and execution structure as learnable decisions, combining iterative tool loading with programmatic tool orchestration to bound context growth and stabilize long-horizon trajectories. Second, we propose rubric-based reinforcement finetuning, which decomposes task success into structured, task-aligned criteria and enables scalable training using small judge models. Across MCP benchmarks, these design choices yield large and consistent gains over generic RL baselines, allowing a 4B SLM to approach frontier-agent performance under far tighter parameter and context budgets.
\end{abstract}

\begin{figure*}[h!]
    \centering
    \includegraphics[width=0.47\textwidth]{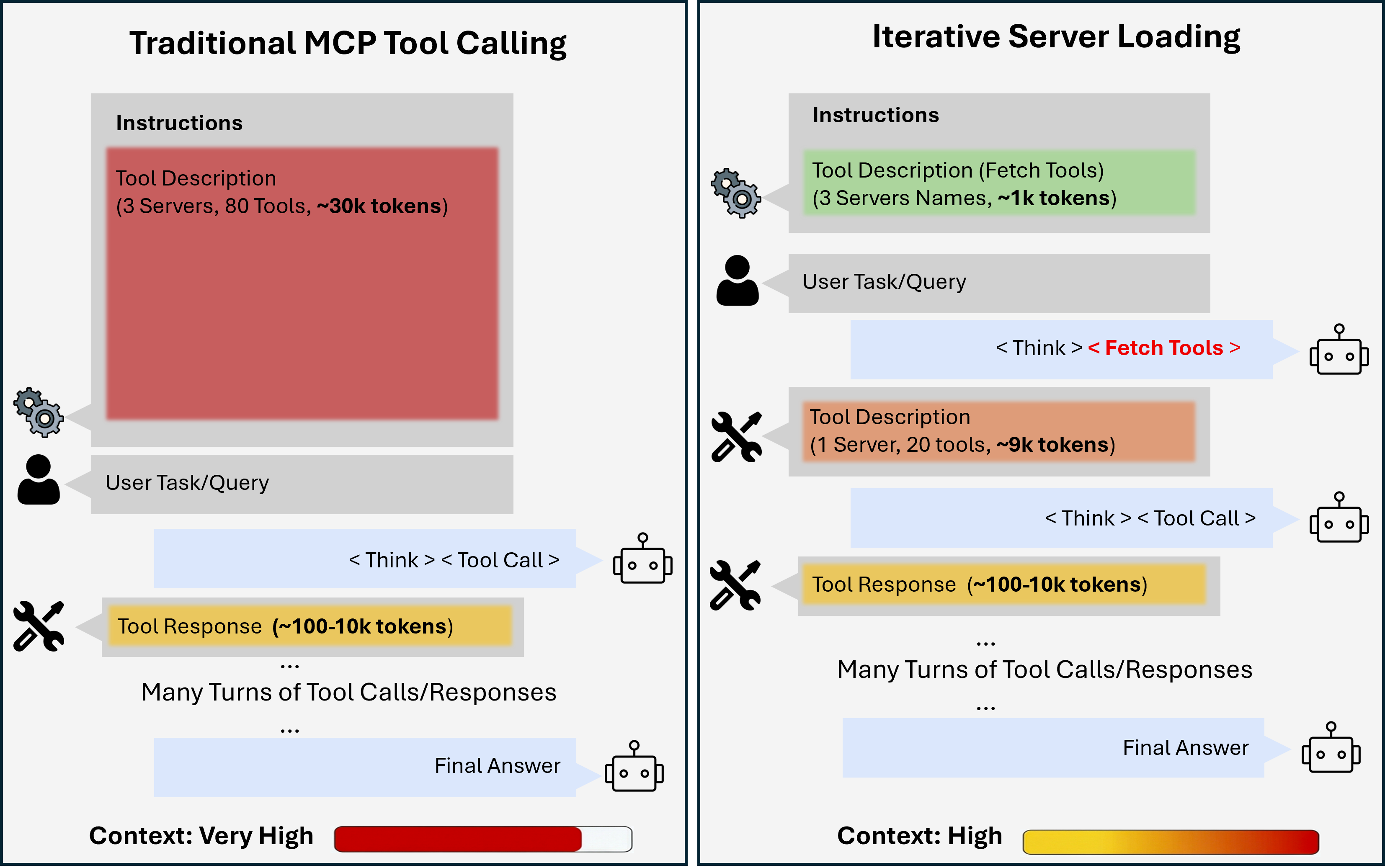}
    \hfill
    \includegraphics[width=0.49\textwidth]{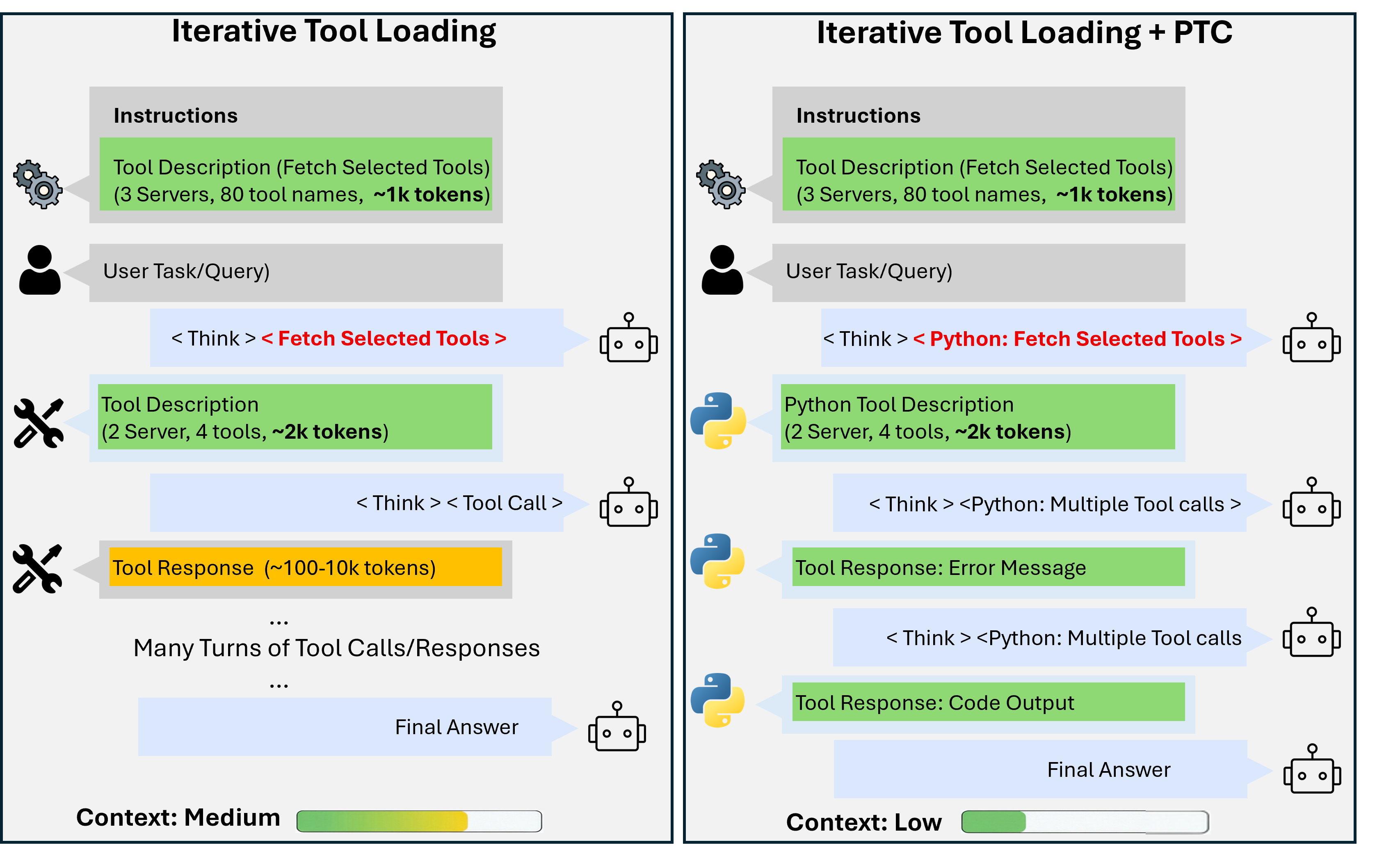}
    \caption{Context growth and execution structure across MCP agent designs.
Traditional MCP agents incur high context costs by loading all tools upfront. ISL and ITL progressively reduce context by scoping server and tool schemas, while ITL+PTC further minimizes prompt growth by moving execution state into programmatic orchestration. }
\vspace{-10pt}
    \label{fig:mcp_comparison}
\end{figure*}
\section{Introduction}
Agentic systems are increasingly deployed in \emph{Model Context Protocol (MCP)} environments, where a single user request may require coordinated planning, reasoning, and execution across multiple external services, tools, and enterprise workflows~\cite{anthropic_mcp}. While MCP’s client--server abstraction enables large-scale integration, it shifts the core challenge from connectivity to \textbf{decision-making at scale}: selecting relevant servers and tools, maintaining state over long trajectories, and operating within tight context and cost budgets~\cite{yao2022react,shinn2023reflexion}. In practice, agents are often connected to hundreds of tools across many MCP servers, and exposing the full tool registry upfront forces reasoning over large, heterogeneous schemas, causing tool definitions and outputs to dominate the context window and degrade efficiency as tool ecosystems scale~\cite{jia2025autotoolefficienttoolselection,cai2023large,wang2024beyond}.

Frontier agent architectures mitigate tool-scale and long-horizon execution through \emph{dynamic tool discovery} and \emph{code-based orchestration}~\cite{anthropic_adv_tool_use_blog_2025}. Rather than preloading all tool schemas, agents retrieve only the servers and tools required at each step, reducing token overhead from verbose definitions and intermediate results. In parallel, \emph{Programmatic Tool Calling (PTC)} replaces JSON-style loops with executable orchestration code, enabling explicit control flow while keeping large artifacts out of the prompt~\cite{anthropic_mcp_code_execution,philschmid_mcp_cli_2026}.

While effective for frontier models, these mechanisms rely on assumptions about context capacity, state robustness, and code reliability that do not hold for \emph{small language models (SLMs)}~\cite{lu2024small,hoffmann2022training}. SLMs are essential for latency-, cost-, and locality-constrained deployments, but naively adopting frontier agent designs exposes structural failure modes~\cite{kim2025towards}. The limitation is not reduced reasoning ability, but a mismatch between the scale assumptions in current architectures and the operating constraints of efficient models.

\textbf{Challenges for SLMs in live MCP environments.}
Agentic systems operating over MCP face a combination of \textit{tool-space explosion}, \textit{long-horizon execution}, and \textit{weak supervision} that disproportionately impacts SLMs~\cite{jia2025autotoolefficienttoolselection}. Agents must reason over large, heterogeneous tool sets distributed across many servers, where dynamic discovery becomes brittle under tight context budgets and costly tool misselection~\cite{cai2023large}. Long-horizon workflows amplify early errors when limited context restricts stable tracking of goals and intermediate state~\cite{shinn2023reflexion}, while code-based orchestration exposes weaknesses in code synthesis, execution, and recovery~\cite{belcak2025small}. Moreover, MCP tasks rarely admit a single verifiable outcome, making outcome-only rewards too sparse for effective credit assignment and leading to brittle behaviors such as premature termination or overuse of familiar tools~\cite{ReinforcementImperfectVerifiers,tang2025beyond}.

Together, these challenges indicate that the primary limitation of SLMs is the absence of mechanisms that explicitly regulate \emph{context growth and execution structure}. This motivates a reframing of agentic reasoning as deciding \emph{what context to acquire, when to acquire it, and how to represent execution compactly}, rather than expanding the prompt~\cite{yao2022react}.

We introduce \pname (\emph{Adaptive Tool Loading and Scoped Context}), a \emph{reinforcement finetuning (RFT)} framework for enabling efficient agentic reasoning with SLMs in large-scale MCP environments. \pname is complementary to frontier agent architectures in that it adopts the same high-level principles of on-demand tool access and structured execution, while explicitly targeting the efficiency-constrained regime where context, computation, and supervision are scarce. Its core novelty lies in treating \emph{context acquisition and execution structure as learnable decisions}, optimized through reinforcement learning rather than fixed architectural choices.

\pname is built around two abstractions. First, \textbf{adaptive tool loading} exposes a compact capability overview and incrementally materializes detailed tool schemas only when required, bounding the tool search space at each step while supporting multi-server workflows. Second, \textbf{orchestration through code} represents long-horizon tool use as executable programs rather than turn-by-turn natural language interactions, enabling compact state representation and reliable control flow without repeatedly surfacing intermediate results to the model.
While these abstractions define how efficient agentic behavior can be represented, learning such behavior remains challenging due to non-verifiable objectives and delayed supervision. Existing RFT approaches for agents typically rely on frontier LLM judges producing synthetic trajectory-level rewards for dimensions such as task success or tool correctness~\cite{ReinforcementImperfectVerifiers,MUARL,guo2025deepseek,artist}. Although effective at small scale, these rewards are coarse, obscure task-specific structure, and require repeated use of expensive frontier judges, limiting scalability.

\pname introduces \textbf{rubric-based reinforcement finetuning}, where an LLM judge evaluates agent behavior using structured, task-level rubrics instead of directly producing a single score. These rubrics decompose evaluation along task-aligned criteria such as completeness, grounding in tool outputs, appropriateness of tool choices, and parameter precision; providing a richer learning signal without step-level annotations. Crucially, once rubrics are generated, \emph{evaluation can be reliably performed by SLMs}, enabling scalable RFT without continued reliance on frontier judges. We further introduce an automated pipeline for rubric generation, allowing task-specific supervision to be produced once and reused throughout training.

We evaluate \pname on a suite of synthetic but realistic MCP tasks constructed using live MCP servers~\cite{wang2025mcpbench}. The training set comprises approximately 300 tasks spanning 28 MCP servers, and the held-out evaluation set includes tasks from both seen and previously unseen servers. We find that task-level rubric rewards yield more stable and effective reinforcement finetuning than generic outcome rewards~\cite{artist,guo2025deepseek}. Notably, under rubric-based evaluation, an open-source \emph{SLM judge} (Qwen3-30B-Instruct) outperforms a GPT-4o-based generic judge and matches stronger frontier judges such as o4-mini. When combined with programmatic orchestration, \pname consistently outperforms JSON-style tool calling in both task completion and efficiency. Our key contributions are 
\begin{enumerate}[itemsep=0mm, parsep=0mm, topsep=0mm, partopsep=0mm]
\item We introduce \pname, a reinforcement finetuning framework that enables efficient agentic behavior through adaptive context control and code-based orchestration for long-horizon tool use. 
\item We show that learning these behaviors is essential: task-level, structured rubric rewards generated by an LLM judge provide stable supervision under non-verifiable objectives without requiring step-level annotations. 
\item We demonstrate that rubric-based evaluation enables SLMs to serve as effective judges, outperforming GPT-4o-based generic judging, enabling scalable and cost-efficient reinforcement finetuning. 
\item We show that task-level rubrics consistently outperform generic outcome rewards, and that code-based orchestration outperforms JSON-style tool calling across diverse MCP tasks. \end{enumerate}

\vspace{-8pt}
\section{\pname Overview}
We present \pname, a framework for training agentic systems to operate efficiently over large and evolving tool spaces under strict context and cost constraints. Rather than eagerly loading all tool schemas, \pname structures agent execution around an inference loop that incrementally materializes only the tool information required at each step.

\pname treats \emph{context acquisition as a decision}: the agent explicitly selects which domains, servers, and tools to load based on the current task state, enabling long-horizon, multi-server workflows without excessive context growth. Execution is represented via \emph{code-based orchestration}, expressing tool use and control flow as compact executable programs instead of turn-by-turn natural language, keeping intermediate state out of the prompt while preserving explicit control flow. These behaviors are learned end-to-end through reinforcement finetuning.
\vspace{-3pt}
\subsection{Iterative Server Loading (ISL)}
ATLAS treats server selection as an explicit decision. At the start of an episode, the agent is given a compact index of available MCP servers and a meta-operation to retrieve tool schemas for a selected server. Conditioned on the task and current execution state, the agent selects a server, materializes only the tools exposed by that server, and executes the required tool calls, loading additional servers incrementally only as needed. This staged exposure avoids eager loading across all servers, preserves context for execution-critical information, and bounds tool selection to a single server at a time, enabling adaptive expansion of the tool space while maintaining a controlled and efficient context footprint.
\vspace{-3pt}
\subsection{Iterative Tool Loading (ITL)}
Iterative server loading enables reasoning over server-level structure but does not ensure grounding in the concrete capabilities of individual tools. In MCP settings, servers often expose hundreds of tools with verbose schemas, making eager materialization impractical and introducing substantial context overhead. \pname addresses this through \textbf{Iterative Tool Loading (ITL)}, which separates high-level planning from detailed tool grounding. Upon loading a server, the agent initially observes only a compact list of tool names, enabling lightweight reasoning about capabilities and plan structure without committing context to full schemas. As execution proceeds, the agent selectively materializes detailed schemas only for the tools required at a given decision point, incrementally grounding the plan as needed. By deferring full tool loading until use, ITL preserves context for execution-critical information while enabling scalable reasoning over large and heterogeneous tool collections.
\vspace{-5pt}
\subsection{Unified Programmatic Tool Orchestration}
Iterative server and tool loading regulate \emph{what} context an agent sees, but conventional JSON-based tool calling still couples reasoning and execution through turn-by-turn natural language interaction. Control flow must be re-derived at each step, intermediate outputs are repeatedly injected into the context, and execution state is tracked only implicitly, leading to rapid context growth and brittle failure modes in long-horizon workflows.

\pname replaces this with \textbf{unified programmatic execution model}, where all tool interactions are mediated by a persistent Python interpreter. Tool calls are expressed as function invocations, control flow is encoded explicitly using programming constructs, and intermediate results are stored in program state rather than surfaced to model, enabling compact, stateful execution without context overhead.

Importantly, this is not equivalent to exposing Python as an additional tool. The interpreter serves as a unified orchestration layer across MCP servers, with execution proceeding via synthesis and refinement of a single program. Errors are handled through localized program edits rather than replaying reasoning traces. Combined with iterative tool loading, this yields disciplined execution in which schemas are materialized only when needed, state remains within the program, and only task-relevant outputs are exposed, which substantially improve robustness and scalability in long-horizon, multi-server settings.
\vspace{-5pt}
\subsection{Scaffolding for Programmatic Tool Orchestration}
MCP tool libraries are out of distribution relative to a model’s coding knowledge, and without additional structure, models often fail to translate plans into correct executable code. Errors such as incorrect function names, mismatched arguments, or improper use of tool outputs can derail long-horizon execution even when high-level reasoning is sound.
\pname introduces a lightweight Python-side scaffolding layer that enables robust programmatic orchestration across previously unseen MCP servers. The scaffold standardizes tool representation, invocation, and debugging, providing a stable execution interface that bridges abstract reasoning and executable behavior.

The scaffold exposes a uniform \texttt{MCPServer} abstraction that presents MCP tools as Python-native functions. Tool schemas from heterogeneous servers are normalized into consistent function signatures, verified against underlying APIs, and bound dynamically, insulating agents from JSON-specific idiosyncrasies. Tool outputs are converted into Python-native data structures and retained in program state rather than re-injected into the model context; when output schemas are missing, representative structures are inferred and reused. To support Iterative Tool Loading, a \texttt{get\_tools\_info()} interface materializes full function signatures and examples only for selected tools. Even when loaded, these Python-native definitions are substantially more token-efficient than raw JSON, enabling precise grounding while preserving context. Additional implementation details and design choices for the scaffold are provided in Appendix~\ref{ptc_scaffold}.


\begin{figure}
    \centering
    \includegraphics[width=0.8\linewidth]{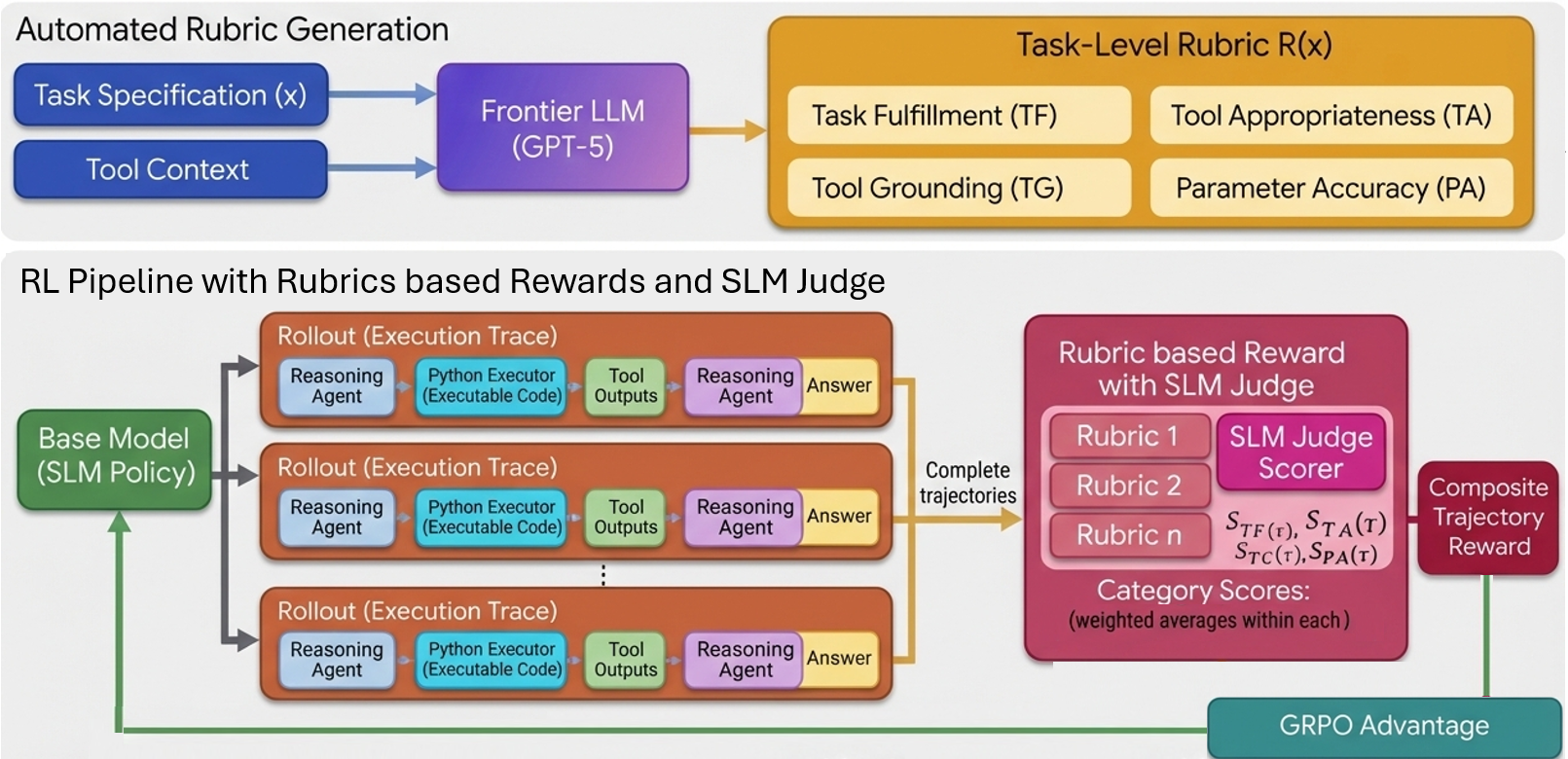}
    \caption{\pname Reinforcement Finetuning approach with Rubrics as Rewards and SLM Judge.}
    \label{fig:placeholder}
    \vspace{-15pt}
\end{figure}

\vspace{-10pt}
\section{Reinforcement Finetuning}

Reinforcement learning for agentic MCP tasks is challenging due to long-horizon, interdependent decision-making. Correct behavior depends on sequences of tool selection, argument construction, execution order, and grounded use of tool outputs, where early errors can invalidate downstream steps despite a plausible final answer. These challenges are compounded by \emph{non-verifiable objectives}: MCP tasks rarely have a single ground truth, and multiple execution trajectories may satisfy user intent, making supervision based on final answers or binary success sparse and often misleading. Effective learning therefore requires reward signals that capture procedural quality and task-specific constraints rather than outcomes alone.


\textbf{Limitations of LLM-as-Judge.}
Most LLM-as-judge frameworks assign a single score, often accompanied by brief reasoning or critique, to an entire agent trajectory~\cite{wang2025mcpbench}. For complex, long-horizon tasks with multiple interacting requirements, this coarse evaluation is insufficient: successful task fulfillment typically depends on satisfying several distinct behavioral and outcome-level criteria. This limitation is particularly acute when LLM-as-judge evaluations are used as rewards in GRPO-style training~\cite{guo2025deepseek}, where group-relative advantage estimation assumes consistent and comparable scoring across trajectories for the same task. In practice, variability in the judge’s internal reasoning and strictness can lead to inconsistent relative rankings, injecting noise into the learning signal and undermining stable policy optimization.

\subsection{Task-Level Rubrics for Agentic RFT}
Agentic MCP tasks are long-horizon and require satisfying multiple interdependent functional criteria. In this regime, generic trajectory-level judgments are difficult for LLM judges to apply consistently, leading to unstable supervision under long contexts and stochastic generation.

We therefore replace direct LLM judge with with LLM judge under the guidance of \textbf{task-level structured rubrics}. For each task, we create a set of rubrics that explicitly define success for each coarse-grained criteria, including task fulfillment, tool appropriateness, grounding in tool outputs, and parameter correctness, together with a weight for each criterion. 

At evaluation time, these fixed rubrics are provided to the LLM judge alongside each trajectory, which assigns scores per rubric. The weighted combination of these scores yields the final reward used for training. By holding rubrics constant across all trajectories for a given task, this approach enforces consistent relative judgments, reduces variance in reward signals, and improves supervision quality compared to unconstrained trajectory-level evaluation. 




\textbf{Automated Rubric Generation.}
Manual rubric design is not scalable in MCP settings due to task diversity and heterogeneous tool usage. \pname therefore generates rubrics automatically \emph{once per task} using a frontier LLM (GPT-5) offline. A rubric is defined as a set of weighted criteria: 
$
\mathcal{R}(x) = \{(C_i, D_i, W_i)\}_{i=1}^{N},
$where $x$ denotes the task specification and $C_i, D_i$ and $W_i$ correspond to criteria name, criteria description and criteria weightage respectively. The rubric generation prompt {(see Appendix~\ref{Rubric generation prompt})} only includes the task specification and available tool context, it is independent of any particular agent trajectory, avoiding bias introduced by long or noisy execution traces.

Each criteria folds under one of four categories capturing core requirements of agentic tool use: task fulfillment, tool appropriateness, grounding in tool outputs, and parameter accuracy. To reduce ambiguity and variance, rubric generation is constrained so that criteria are (i) observable from execution traces, (ii) non-overlapping, and (iii) aligned with functional task requirements rather than surface-level language quality.
\vspace{-5pt}
\paragraph{RFT with Rubric-Based Rewards.}
We adopt a cold-start reinforcement finetuning setup in which all supervision is provided by rubric-based rewards, without step-level annotations or supervised warm-start. For each task $x$, a task-specific rubric $\mathcal{R}(x)$ is generated offline using GPT-5 and fixed throughout training. All rollouts for the same task are evaluated against the same rubric, enabling stable relative comparisons during advantage estimation under GRPO.

Each rubric criteria is scored, and then it's weighted sums are normalized under the category it folds into. There are four categories for trajectory quality:
(i) \textbf{Task Fulfillment (TF)};
(ii) \textbf{Tool Appropriateness (TA)};
(iii) \textbf{Tool Grounding (TG)};
(iv) \textbf{Parameter Accuracy (PA)}.

For a trajectory $\tau$, an LLM-based judge assigns a score $d_i(\tau) \in [0,1]$ to each rubric criterion $C_i$, with $1$ indicating full satisfaction. Each rubric criterion $C_i$ is associated with a weight $W_i$. For each category $R \in \{\mathrm{TF}, \mathrm{TA}, \mathrm{TG}, \mathrm{PA}\}$ containing $N_R$ criteria, we compute a weighted category score:
$
S_R(\tau)
=
\frac{\sum_{i=1}^{N_R} W_i\, d_i(\tau)}
     {\sum_{i=1}^{N_R} W_i}.
$
The final trajectory reward is a \emph{composite reward} obtained by aggregating category scores:
$
R(\tau)
=
\sum_{R \in \{\mathrm{TF}, \mathrm{TA}, \mathrm{TG}, \mathrm{PA}\}}
\alpha_R \, S_R(\tau),
$
with fixed category weights $\alpha_R$ shared across all rollouts for the task, assigning higher weight to task fulfillment. Examples of the automated rubrics generation are provided in Appendix~\ref{sec:automated_rubrics}.

\vspace{-5pt}
\paragraph{Scaling Judging from Frontier LLMs to SLMs.}
A key advantage of rubric-based rewards is scalability. In prior LLM-as-judge approaches, evaluating long trajectories with frontier models often dominates training cost. Rubric-conditioned scoring reduces judging from holistic trajectory assessment to evaluating concrete, well-specified criteria, substantially lowering the reasoning burden on the judge.

As a result, small language models (SLMs) can serve as effective judges when guided by task-level rubrics. We find that SLM judges provide stronger and more stable learning signals than frontier LLM judges using generic rewards, and achieve better performance than stronger frontier judges (e.g., GPT4o) under rubric-based evaluation. This enables scalable and cost-efficient reinforcement finetuning for long-horizon, non-verifiable agentic MCP tasks.

During optimization, gradients are masked over tool-output tokens so that learning applies only to model-generated content. This focuses policy updates on planning, tool selection, argument construction, execution control, and termination decisions, rather than imitation of tool responses.

\vspace{-10pt}
\section{Experimental Setup}
\subsection{MCP Tasks and Environments}
We evaluate \pname on synthetic but realistic agentic tasks over live MCP servers, requiring long-horizon reasoning, tool discovery and invocation, and grounding in tool outputs. Tasks are non-verifiable, admitting multiple valid execution trajectories.

\textbf{Train data.} Our task data generation is derived from MCPBench~\cite{wang2025mcpbench}, from which we select 28 MCP servers spanning heterogeneous tool ecosystems (search, structured data, file systems, and enterprise APIs). Using MCPBench’s synthetic task proposal pipeline, we generate over 1,000 multi-server tasks with 1–3 servers per task and stratified difficulty (Easy, Medium, Hard), corresponding to approximately 2–3, 4–5, and 6+ distinct requirements. Tasks are generated using \texttt{o4-mini} and filtered with automated judges to retain high-quality tasks (solvability $>9$, utility $>8$).

Since many generated tasks are either unsolvable with the available tools or exceed the practical limits of SLMs, we further filter tasks by executing them with Kimi K2 Thinking (1T parameters) and retain only those achieving a task fulfillment score of at least 4/10, yielding a final set of 304 solvable, high-quality tasks (see Table~\ref{tab:task_distribution} in Appendix~\ref{app:splits}).

\textbf{Test data.} We evaluate generalization on two held-out test sets. The first consists of 104 unseen tasks from the MCPBench benchmark drawn from previously seen 28 servers, measuring task-level generalization. The second, \pname-Test, comprises 100 tasks drawn from 11 unseen MCP servers + 28 MCP servers, evaluating server-level and tool-level generalization under distribution shift. These tasks are designed to assess generalization to unseen servers and novel tool combinations. Appendix~\ref{app:dataset} presents the details of various servers used in MCPBench and \pname-Test along with task distribution.

\subsection{Models}
We evaluate \pname on open-weight language models spanning different parameter scales and levels of agentic capability, (i) \textbf{Qwen2.5-7B-Instruct}~\cite{Yang2024Qwen2_5} (7B), chosen for strong instruction-following and suitability for reinforcement finetuning and (ii) \textbf{Qwen3-4B-Instruct-2507}~\cite{Yang2025Qwen3} (4B), a smaller model with native tool-calling support. We use 32K context length for both models.

As a strong frontier baseline, we additionally evaluate \textbf{Kimi K2 Thinking}~\cite{team2025kimi} (1T parameters, 80K Context), configured with traditional MCP execution in which all available tool schemas are eagerly loaded into the system prompt. We further evaluate Kimi under our execution scaffolding with \textbf{Iterative Server Loading (ISL)} and \textbf{Iterative Tool Loading (ITL)} to isolate the impact of context control and execution structure independent of model scale.

\subsection{Evaluation Metrics}
We evaluate models performance on MCP tasks using an \textbf{LLM-as-judge} framework. We adopt the evaluation prompt provided by \texttt{MCPBench}~\cite{wang2025mcpbench}, which scores complete agent trajectories given the task specification and requirements. The judge evaluates performance across four categories: \textbf{Task Fulfillment (TF)}—whether core task requirements are satisfied; \textbf{Tool Appropriateness (TA)}—whether selected tools are relevant and necessary; \textbf{Tool Grounding (TG)}—whether tool outputs are used faithfully and correctly; and \textbf{Parameter Accuracy (PA)}—the correctness and precision of tool arguments. Following \texttt{MCPBench}, we use \texttt{o4-mini} as the LLM judge for its robustness in validation. The full evaluation prompt is provided in the Appendix~\ref{app:Base judge prompt}.

\subsection{Reinforcement Finetuning Setup}


Task-level rubrics are generated once per task using a frontier model (GPT-5)~\cite{gpt5}, while trajectory evaluation during training is performed by an \textbf{SLM judge} (Qwen3-30B-Instruct)~\cite{Qwen3-30B} conditioned on these rubrics. Generic LLM Judge reward baselines are evaluated using frontier judges (GPT-4o)~\cite{gpt4o}.

We compare agent variants differing only in execution and learning mechanisms: (i) traditional MCP execution with eager tool loading, (ii) iterative server loading (ISL), (iii) iterative server and tool loading (ISL+ITL), and (iv) \pname (ISL+ITL with programmatic orchestration).

\textbf{Implementation Details.}
All experiments are conducted on machines equipped with $8\times$ NVIDIA B200 GPUs. We use the \texttt{verl} library as the reinforcement learning framework and extend it to support MCP tool calling, including a custom rollout structure, truncation of overlong tool responses, and integration of LLM-based judge evaluation into the training loop.
Reward computation and policy updates are performed online during training. All models share fixed hyperparameters and identical optimization settings to ensure fair comparison. Additional implementation details are provided in Appendix~\ref{app:Training Hyperparameters}.

\section{Results}

\definecolor{basegray}{RGB}{245,245,245}
\definecolor{frontierblue}{RGB}{232,240,254}
\definecolor{slmgreen}{RGB}{232,247,236}
\definecolor{ptcorange}{RGB}{255,243,230}

\begin{table}[t]
\centering
\footnotesize
\renewcommand{\arraystretch}{0.3}
\setlength{\tabcolsep}{1pt}
\resizebox{0.7\columnwidth}{!}{
\begin{tabular}{clccc}
\toprule
\textbf{\#} &
\textbf{Variant} &
\makecell{\textbf{MCPBench} \\ \textbf{Task Ful.}} &
\makecell{\textbf{Avg} \\ \textbf{Turns}} &
\makecell{\textbf{Avg} \\ \textbf{Tokens}} \\
\midrule

\multicolumn{5}{l}{\textbf{Kimi-K2 Thinking (Frontier Baseline)}} \\
\midrule
\rowcolor{basegray}
1  & All Tools Loaded              & 4.38 &  20 & 23768 \\
\rowcolor{basegray}
2  & Iterative Server Loading      & 4.11 &  27 & 21747 \\
\rowcolor{basegray}
3  & Iterative Tool Loading        & 3.62    & 28 & 18290 \\
\midrule
	
\multicolumn{5}{l}{\textbf{Qwen2.5-7B-Instruct (Server Loading)}} \\
\midrule
\rowcolor{basegray}
4  & Iterative Server Loading      & 2.33  & 25 & 11192 \\
\rowcolor{frontierblue}
5  & + RL w/ GPT-4o Judge (Generic)          & 3.02  & 19 & 10307 \\
\rowcolor{slmgreen}
6  & + RL w/ Qwen3-30B Judge (Generic)       & 2.84 & 24 & 12201 \\
\rowcolor{frontierblue}
7  & + RL w/ GPT-4o Judge (Rubrics)          & 3.13  & 19 & 11882 \\
\rowcolor{slmgreen}
8  & + RL w/ Qwen3-30B Judge (Rubrics)       & \textbf{3.18}  & 21 & 13075 \\
\midrule

\multicolumn{5}{l}{\textbf{Qwen3-4B-Instruct (Server Loading)}} \\
\midrule
\rowcolor{basegray}
9  & Iterative Server Loading      & 2.73  & 24	 & 9152 \\
\rowcolor{frontierblue}
10 & + RL w/ GPT-4o Judge (Generic)           & 3.25  & 17 & 9372 \\
\rowcolor{slmgreen}
11 & + RL w/ Qwen3-30B Judge (Generic)        & 3.12  & 21 & 11530 \\
\rowcolor{frontierblue}
12 & + RL w/ GPT-4o Judge (Rubrics)           & 3.43  & 18 & 10654 \\
\rowcolor{slmgreen}
13 & + RL w/ Qwen3-30B Judge (Rubrics)        & \textbf{3.87} &  19 & 11142 \\
\midrule

\multicolumn{5}{l}{\textbf{Qwen3-4B-Instruct (Tool Loading)}} \\
\midrule
\rowcolor{basegray}
14 & Iterative Tool Loading                & 2.36  & 20 & 9045 \\
\rowcolor{ptcorange}
15 & + PTC                                 & 2.94  & 18 & 13462 \\
\rowcolor{slmgreen}
16 & + RL w/ Qwen3-30B J. (Generic) & 3.48  & 23 & 12815 \\
\rowcolor{ptcorange}
17 & + PTC + RL w/ Qwen3-30B J. (Generic)  & 3.91 & 20 & 12951 \\
\rowcolor{slmgreen}
18 & + RL w/ Qwen3-30B J. (Rubrics) & 4.03  & 20 & 11151 \\
\rowcolor{ptcorange}
19 & + PTC + RL w/ Qwen3-30B J. (Rubrics)  & \textbf{4.15}& 18 & 13400 \\
\midrule

\end{tabular}
}
\caption{
Performance comparison across model families, execution strategies, and reinforcement finetuning configurations.
\textbf{Gray}: base (no learning);
\textbf{Blue}: frontier judge (GPT-4o);
\textbf{Green}: SLM judge (Qwen3-30B);
\textbf{Orange}: Programmatic Tool Calling (PTC).
Task Fulfillment (TF) is reported on a 0--10 scale. 
}
\vspace{-18pt}
\label{tab:model_comparison}
\end{table}

\begin{figure*}
    \centering
    \includegraphics[width=0.36\linewidth]{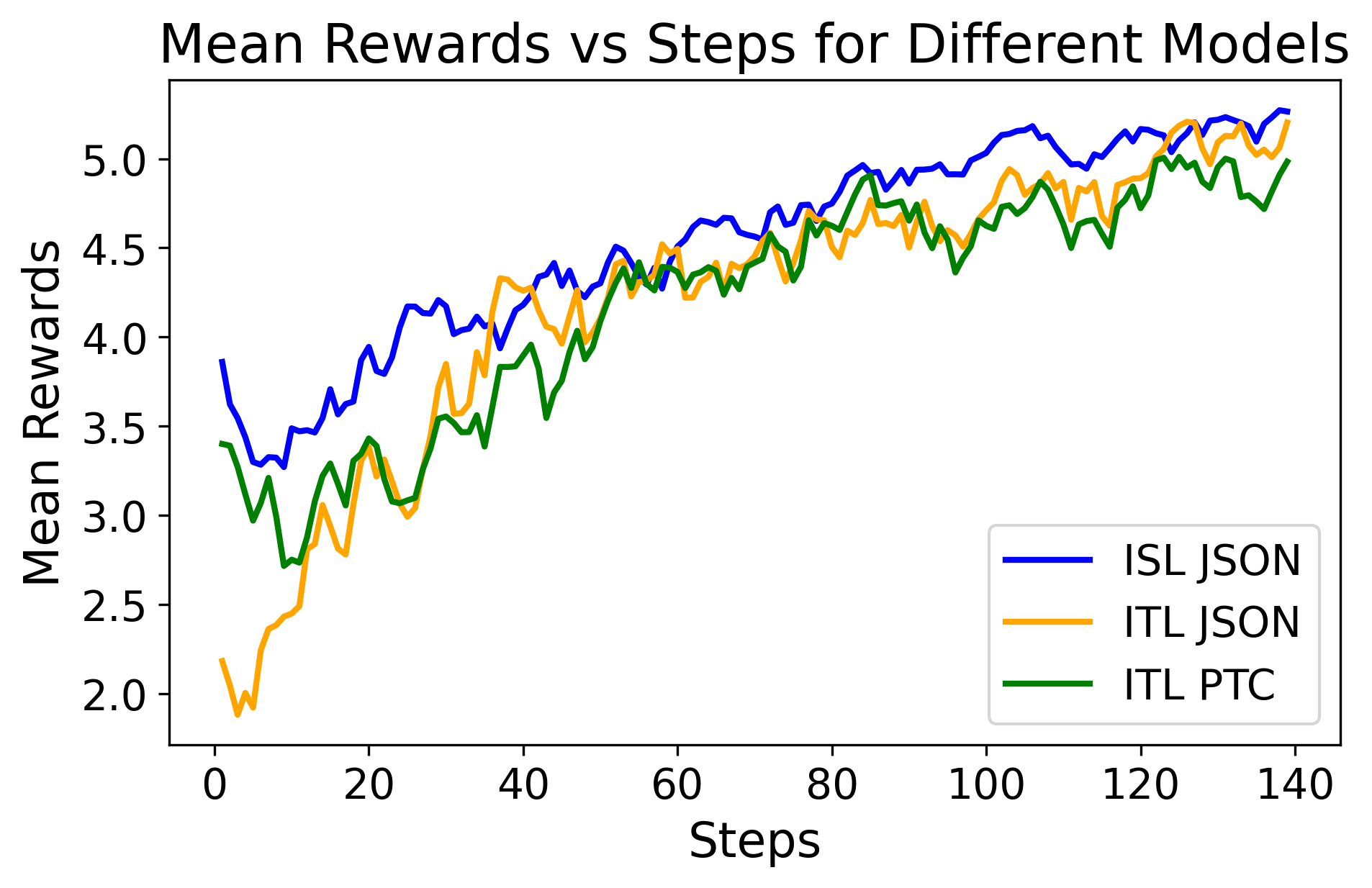}
    \includegraphics[width=0.36\linewidth]{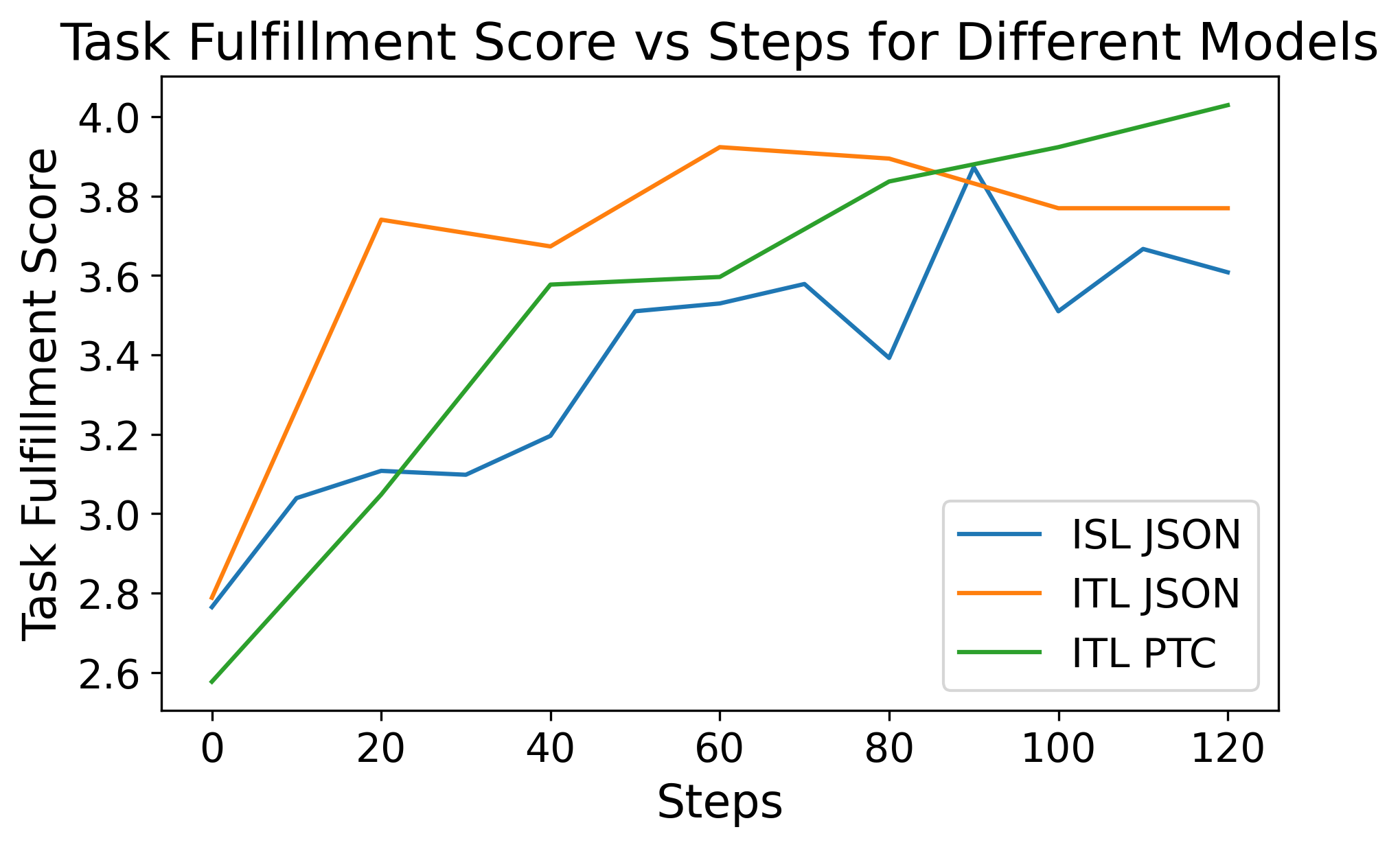}
    \vspace{-8pt}
    \caption{\pname reinforcement finetuning training and validation curves under ISL, ITL, and ITL+PTC, showing composite training rewards (left) and task-fulfillment-based validation performance (right).}
    \label{fig:train_graph}
    \vspace{-10pt}
\end{figure*}

\subsection{Overall Performance on MCPBench}
Table~\ref{tab:model_comparison} reports \textbf{Task Fulfillment (TF)} on MCPBench across model families, execution strategies, and reinforcement finetuning configurations. TF (0--10) is our primary metric and directly measures end-to-end success on compositional, long-horizon MCP tasks. Additional diagnostic metrics and \pname-Test results are reported in Appendix~\ref{app:Detailed Results}, Table~\ref{tab:model_comparison_appendix} and~\ref{tab:model_comparison_appendix_atlas}.
\vspace{-3pt}
\begin{center}
\colorbox{green!8}{
\parbox{0.95\linewidth}{
\textbf{Key takeaway.} Structured execution alone is insufficient for small models, but when combined with reinforcement finetuning—especially rubric-based supervision—efficient SLMs approach frontier-level performance.
}}
\end{center}\vspace{-3pt}

Across all model scales, base agents without learning perform poorly, typically achieving TF in the \textbf{2--3} range (Qwen2.5-7B ISL: \textbf{2.33}, Row~4; Qwen3-4B ISL: \textbf{2.73}, Row~9; Qwen3-4B ITL: \textbf{2.36}, Row~14). Introducing reinforcement finetuning yields large and consistent gains, while structured execution amplifies these improvements by stabilizing long-horizon trajectories.

In contrast, the frontier model \textbf{Kimi-K2 Thinking} achieves strong performance even with all tools eagerly loaded (TF \textbf{4.38}, Row~1), reflecting its ability to tolerate large context windows. Importantly, Kimi does \emph{not} benefit from ITL (TF drops to \textbf{3.62}, Row~3), as it is not trained to reason over JSON-restricted tool views. A similar pattern appears for small models under cold-start ITL.
\vspace{-10pt}
\subsection{Effect of Execution Structure}
We first isolate the impact of execution structure under cold-start conditions, holding learning fixed.

\textbf{Iterative Server Loading (ISL).}
Under ISL, both SLMs exhibit low TF (Qwen2.5-7B: \textbf{2.33}, Row~4; Qwen3-4B: \textbf{2.73}, Row~9), confirming that server-level context control alone is insufficient for non-verifiable MCP tasks.

\textbf{Iterative Tool Loading (ITL).}
Moving from ISL to ITL reduces schema exposure and token usage (Qwen3-4B: \textbf{9{,}152} $\rightarrow$ \textbf{9{,}045} tokens, Rows~9 $\rightarrow$ 14), but yields only modest gains in TF (\textbf{2.73} $\rightarrow$ \textbf{2.36}). This mirrors the behavior seen in Kimi (Rows~2--3): models not explicitly trained to reason over JSON-limited tool views struggle to fully exploit ITL under cold-start conditions.

\textbf{Programmatic Tool Calling (PTC).}
Adding \textbf{PTC} under ITL improves execution robustness even without learning. For Qwen3-4B, TF increases from \textbf{2.36} to \textbf{2.94} (Rows~14 $\rightarrow$ 15), despite higher token usage due to executable code. This indicates that programmatic control flow reduces execution failures and stabilizes long-horizon behavior, providing a stronger substrate for learning.
\vspace{-3pt}
\begin{center}
\colorbox{green!8}{
\parbox{0.95\linewidth}{
\textbf{Key takeaway.} Without learning, ITL yields modest gains ($\sim$10\%), while PTC delivers stronger improvements ($\sim$25\%) by stabilizing execution—showing that structure alone is insufficient.
}}
\end{center}\vspace{-10pt}
\subsection{Impact of Reinforcement Finetuning}
We now examine the effect of reinforcement finetuning (RFT), holding execution structure fixed.

\textbf{ISL + RFT.}
Under ISL, RFT produces large gains across both model scales. For Qwen2.5-7B, TF improves from \textbf{2.33} (Row~4) to \textbf{3.02} with a generic frontier judge (Row~5), and further to \textbf{3.18} with rubric-based supervision and an SLM judge (Row~8). Qwen3-4B exhibits the same pattern, improving from \textbf{2.73} (Row~9) to \textbf{3.25} (Row~10) and up to \textbf{3.87} with rubric-based RFT (Row~13).

\textbf{ITL + RFT.}
Under ITL, generic RFT substantially improves performance (Qwen3-4B: \textbf{2.36} $\rightarrow$ \textbf{3.48}, Rows~14 $\rightarrow$ 16). However, JSON-based ITL remains less effective than its PTC counterpart, even after learning.

\textbf{ITL + PTC + RFT (Best-performing SLM regime).}
The strongest results are obtained when \textbf{ITL, PTC, and RFT are combined}. For Qwen3-4B, TF reaches \textbf{3.91} under generic RFT with an SLM judge (Row~17), outperforming JSON-based ITL variants. While we do not yet apply rubric-based rewards to PTC, due to the challenge of defining concrete rubrics over executable code. This configuration already represents the best-performing SLM agent.
\vspace{-3pt}
\begin{center}
\colorbox{green!8}{
\parbox{0.95\linewidth}{
\textbf{Key takeaway.} RFT delivers the largest gains (\textbf{+35--65\%}), with ITL+PTC providing the strongest execution substrate for learning.
}}
\end{center}\vspace{-3pt}
\vspace{-10pt}
\subsection{Rubric-Based Rewards and Judge Scaling}
We analyze the effect of reward design and judge choice.

\textbf{Rubrics vs.\ generic rewards.}
Across both model scales and execution strategies, \textbf{rubric-based RFT consistently outperforms generic scalar rewards}. For Qwen2.5-7B under ISL, TF improves from \textbf{3.02} to \textbf{3.18} (Rows~5 $\rightarrow$ 8). For Qwen3-4B, the gains are larger: \textbf{3.25} $\rightarrow$ \textbf{3.87} (Rows~10 $\rightarrow$ 13). These results highlight the importance of task-aligned credit assignment in non-verifiable MCP tasks.

\textbf{Frontier vs.\ SLM judges.}
Under generic rewards, GPT-4o and Qwen3-30B perform comparably (e.g., Qwen2.5-7B: Rows~5 vs.~6). Under rubric-based supervision, however, the \textbf{SLM judge consistently outperforms the frontier judge} (Qwen3-4B: \textbf{3.87} vs.\ \textbf{3.43}, Rows~13 vs.~12). Since all methods are evaluated using a fixed external judge (o4-mini), these gains reflect improved training dynamics rather than evaluation bias.
\vspace{-3pt}
\begin{center}
\colorbox{green!8}{
\parbox{0.95\linewidth}{
\textbf{Key takeaway.} Rubric-based supervision improves task success by up to \textbf{20\%} and allows SLM judges to outperform frontier judges.
}}
\end{center}\vspace{-3pt}
\textbf{Interaction B/W Execution Structure and Learning}
A central finding is the strong \textbf{positive interaction} between execution structure and learning. Structured execution without learning yields limited gains (Rows~14--15), while RFT under weak structure saturates early (Rows~5--8, 10--13). In contrast, combining disciplined execution (ITL+PTC) with RFT produces the highest TF, indicating that execution structure enables effective credit assignment over long horizons rather than acting as a standalone improvement.

\textbf{Efficiency and Training Dynamics}
ISL and ITL significantly reduce token usage by constraining schema exposure (Qwen3-4B: \textbf{12{,}906} tokens under ISL, Row~9, vs.\ \textbf{9{,}045} under ITL, Row~14). PTC increases token usage due to executable representations (Row~15), but reduces interaction turns and enables higher TF once learning is applied (Rows~16--17). The added cost reflects explicit execution structure and is offset by improved task success.

Figure~\ref{fig:train_graph} shows that PTC and ITL achieve slightly lower training rewards across steps, yet outperform other variants on the target task fulfillment (TF) metric at test time. This divergence indicates stronger generalization to unseen tasks and suggests that these methods produce trajectories that are less susceptible to reward hacking the training-time judge, prioritizing true task success over optimizing the train reward.

\subsection{Closing the Gap to Frontier Models}
The strongest SLM configuration—\textbf{Qwen3-4B with ITL, PTC, and RFT with Rubrics and SLM Judge}—achieves \textbf{4.15/10 TF} (Row~19), approaching the frontier baseline \textbf{Kimi-K2 Thinking} at \textbf{4.38/10 TF} (Row~1), despite operating with \textbf{orders of magnitude fewer parameters} and \textbf{far tighter context budgets}. This also suggests that the proposed techniques are complementary, yielding consistent additive improvements when applied together.

\vspace{-3pt}
\begin{center}
\colorbox{green!8}{
\parbox{0.95\linewidth}{
\textbf{Key takeaway.} With disciplined execution structure and RFT, a 4B SLM approaches the performance of a frontier Kimi-K2-Thinking agent despite operating under far tighter model and context constraints.
}}
\end{center} 

\vspace{-10pt}
\section{Related Work}
\label{sec:related}
Large language models have increasingly evolved from static generators into interactive agents capable of multi-step reasoning, tool use, and environment interaction. \pname is most closely related to work on learned tool invocation, reinforcement learning for agentic behavior, adaptive context and tool selection, and programmatic execution.

\textbf{Learned Tool Invocation.}
Early work established mechanisms for augmenting LLMs with external tools. Toolformer demonstrated self-supervised API call insertion to improve downstream performance \cite{schick2023toolformer}. More recent approaches treat tool calls as executable code, enabling stateful, programmatic interaction and stronger control over long-horizon tasks \cite{wang2024codeact}.

\textbf{Reinforcement Learning for Agentic Tool Use.}
Reinforcement learning has been widely used to teach agents when and how to invoke tools. ReTool applies outcome-based RL to interleave reasoning with real-time code execution without step-level supervision \cite{feng2025retool}. ARTIST unifies reasoning, tool selection, and environment interaction within an RL framework for autonomous tool use \cite{artist}. Related work explores learning stability and credit assignment, including turn-level rewards \cite{zeng2025turnlevel} and modeling interactive user dynamics \cite{MUARL}.

\textbf{Dynamic Tool Selection and Context Control.}
As static tool inventories become impractical, several approaches focus on adaptive tool selection. AutoTool trains models to dynamically select tools using large-scale rationale data and dual-phase optimization \cite{jia2025autotoolefficienttoolselection}. ToolSample combines curriculum learning with dynamic sampling \cite{feng2025toolsample}, while Tool-Star coordinates multiple tools under RL supervision \cite{dong2025toolstar}, highlighting the importance of structured training for scalable tool use.
MCP-Bench evaluates tool-using agents in Model Context Protocol environments, exposing challenges from multi-server tool ecosystems and large context footprints \cite{wang2025mcpbench}.

\textbf{Programmatic Execution.}
Programmatic orchestration has emerged as an alternative to turn-by-turn natural language tool calls. Executable action frameworks reduce context overhead and enable stateful, verifiable control flow \cite{wang2024codeact}. This paradigm underlies practical systems such as Anthropic’s programmatic tool calling interfaces \cite{anthropic_ptc_docs_2025,anthropic_adv_tool_use_blog_2025}, MCP tooling like \texttt{mcp-cli} \cite{philschmid_mcp_cli_2026}, and terminal-based agent environments such as SETA \cite{camel_seta_blog_2026}.

\textbf{Evaluation and Supervision.}
Sparse rewards remain a major challenge for RL in complex agentic settings. Recent work shows that structured, rubric-based evaluation provides denser supervision and improves training stability in non-verifiable domains \cite{yu2025demystifying}.

\textbf{\pname.}
Prior work addresses individual aspects of agentic reasoning—tool invocation, reinforcement learning, dynamic selection, or programmatic execution. \pname differs by jointly treating \emph{context acquisition as a learned decision} and \emph{execution structure as a first-class optimization target} for efficiency-constrained agents operating over large MCP tool spaces. By combining adaptive tool loading, compact programmatic orchestration, and task-level rubric-based reinforcement finetuning, \pname provides an integrated approach not explored in prior work.

\newpage
\section{Conclusion}
This work argues that the central challenge in agentic MCP systems is not model scale, but how agents acquire context, execute actions, and learn from weak supervision. \pname addresses this by learning context control and execution structure through adaptive tool loading, programmatic orchestration, and task-level rubric supervision. Empirically, execution structure alone improves efficiency but yields limited task success; substantial gains arise only when reinforcement finetuning provides task-aligned credit assignment. Structured rubrics are central to this effect, enabling small judge models to match or surpass frontier judges and making training scalable and cost-efficient. As a result, a 4B SLM recovers nearly 90\% of frontier-agent performance despite operating under far tighter parameter and context budgets. These results mark a shift from scale-driven to structure- and learning-driven agent design.

\bibliography{example_paper}
\bibliographystyle{iclr2026_conference}

\appendix
\section*{Appendix}

\section{Programmatic Tool Calling Implementation}
\label{ptc_scaffold}

A key piece of the puzzle to make PTC work is coming up with a scaffold atop the python executor that makes it possible for the SLM to go from an initial plan or pseudocode to a working code given that our MCPServer library and its tools are out-of-distribution for the model's coding knowledge.

While trying to write working code from the pseudocode/plan using unseen MCP Server function formats, the model can make three following types of mistakes:
\begin{enumerate}[noitemsep, topsep=0pt]
    \item Incorrect function names
    \item Incorrect argument names and formats
    \item Incorrect output access and composition.
\end{enumerate}
To address these issues, we implement 1. Input schema normalization, 2. MCPServer class abstraction and dynamic function allocation, 3. Server output conversion 4. Function output schema and examples fetching 5. Iterative Tool Loading Implemetation and 6. Informative error logs and hints.

\subsection{
Input schema normalization
}

Different MCP servers tend to reveal different levels of information about their tools and may even use different JSON keys to describe similar properties. We run an LLM script to map these tool input schemas into a common, clean schema without losing any information. We then process these JSON schemas into Python function and argument names. This process is run in parallel for all tools and servers, and the results are stored offline. The fetched function names and argument names are programmatically verified to match the original JSON schemas.

Example python converted schema: 

{\footnotesize
\texttt{home\_manager\_search(query: str (required), limit: int (optional, default=20, max=100))}
}

Note that most MCP servers do not define the output schema, and at this stage, we also do not have the Python output formats.

\subsection{MCPServer class abstraction and dynamic function allocation}

The \texttt{MCPServer} class interfaces between python native code and the MCP servers' json API. The class is instantiated at a server level (e.g.\ \texttt{time\_mcp = MCPServer("Time MCP")}). This class dynamically sets the attributes of the class based on the \texttt{server\_name} its initialized with(e.g. \texttt{time\_mcp.get\_current\_time(timezone='America/New\_York')})
and then maps them to the MCP server calls. For a specific server, all the tool names and the args are fetched from the saved mapping created in the previous step.

\subsection{Server output conversion}

The responses from the MCP servers always come as serialized text; we need to load them into appropriate python types so that they can be used for sequential processing in code. We use \texttt{ast.literal\_eval()} for this conversion which follows a "top-down" approach, moving from structural complexity (lists/tuples/dicts) to primitive data types (int/float/str).

\subsection{Function output schema and examples fetching}

Once we have the python function calling wrapper around the MCP servers' API, we can get the output schemas. We design a simple tool calling loop with an LLM endpoint and prompt it to execute the python functions with with realistic arguments given the descriptions. The LLM calls the tools till it succeeds in getting the outputs, then it stores the output format and an example for each tool in the same python native format. 

\subsection{Iterative Tool Loading Implemetation}
To enable orchestration across tool calls and servers, we need to give the model a good understanding of the exact tool usage. To achieve this, we implement a \texttt{mcp\_server\_obj.get\_tools\_info([])} function that fetches the full tool schemas and examples for the requested tool names from the pre-saved info in the last step. Note that even with the full schemas and examples, python definitions take a lot less tokens that raw json schemas.

\subsection{Informative error logs and hints}
The logic for most agentic tasks is simple enough for the model to get. The challenge is in writing the code using a completely unseen library without taking too many trial-and-error steps. We nudge the model towards correcting itself in the following ways:

\begin{itemize}
    \item Incorrect function names: Instead of a plain "object of class MCPServer has no attribute 'incorrect\_func\_name'", we give the message " MCP Server 'server\_name' doesn't have the tool 'incorrect\_func\_name" available tools : {'list of tool names'}, did you mean {closest tool}?"
    \item Argument names and formats: In this case, most of the servers themselves give a clear feedback on the arg errors, we log the same for every failed function call in the code block.
    \item Incorrect output access: If the model encounters an error while using the returned value from any server as a variable, we pass its output schema in the logs. We also provide hints like "You have tried to access a string as a dict, please the output logs or use \texttt{get\_tools\_details[]}" instead of a plain "str indexes must be integers".
\end{itemize}

\subsection{Example ISL vs ITL+PTC Behavior}
The Qwen3-4B ISL agent with json tool calling examples shows multiple tool calls and verbose intermediate reasoning when the steps have dependencies between them:
\begin{tcolorbox}[breakable, colback=gray!5, colframe=black, title= ISL Example 1]

[user]\\
Task: I’m pulling together a report on last quarter’s harvest from our 10 farms, and honestly I need some hard numbers. We recorded yields of 120, 150, 150, 200, 180, 170, 160, 140, 130, and 155 tons. \\

Here’s what I’m trying to nail down:
- What’s our total output, average yield per farm, the median and the most common harvest size, plus our lowest and highest yields and the overall spread?\\
- Then, at \$30 a ton, what does that translate to in revenue?\\
- After covering \$2,000 in fixed costs per farm (so 10 farms total), what’s left as net profit and what’s our profit margin when you express it as a percentage (rounded to the nearest whole number)?\\
- Finally, I’m curious about the gap between our top-performing farm (200 tons) and the average yield—if that difference is more than 30 tons, I want to budget extra fertilizer at \$10 per ton of that gap (and round up); if it’s 30 or less, I’ll stick with a \$500 allowance (and round down).\\

Could you crunch all those figures? I really need solid data—can’t go to my boss with just guesses. Thanks!\\

Please ensure all findings are supported by concrete data and verifiable sources. I need specific numbers and evidence, not generalizations..\\

The following servers are available:\\
Time MCP\\
Math MCP\\

[assistant]
\begin{verbatim}
<think>
First, I need to determine which server can provide the necessary 
calculations.
The Math MCP server seems to be the most relevant for this task as it can
handle statistical and numerical operations.
</think>

<tool_call>
{"name": "fetch_tools", "arguments": {"server": "Math MCP"}}
</tool_call>
\end{verbatim}\\

[user]
\begin{verbatim}
<tool_response>
Tools in server='Math MCP'

Tool: `Math MCP:add` (Server: Math MCP)
  Description: Adds two numbers together
  Input Schema:
{
  "type": "object",
  "properties": {
    "firstNumber": {
      "type": "number",
      "description": "The first addend"
    },
    "secondNumber": {
      "type": "number",
      "description": "The second addend"
    }
  },
  "required": [
    "firstNumber",
    "secondNumber"
  ],
  "additionalProperties": false
}
.....
</tool_response>
\end{verbatim}

[assistant]
\begin{verbatim}
<tool_call>
{"name": "Math MCP:sum", "arguments": {"numbers": [120, 150, 150, 200, 
180, 170, 160, 140, 130, 155]}}
</tool_call>
\end{verbatim}

[user]
\begin{verbatim}
<tool_response>
1555
</tool_response>
\end{verbatim}

\ldots

[assistant]
\begin{verbatim}
<tool_call>
{"name": "Math MCP:median", "arguments": {"numbers": [120, 150, 150, 200, 
180, 170, 160, 140, 130, 155]}}
</tool_call>
\end{verbatim}

[user]
\begin{verbatim}
<tool_response>
undefined
</tool_response>
\end{verbatim}

[assistant]
It seems the \texttt{median} function did not return a value, which might be due to the way the function is implemented or the specific version of the tool. Let's calculate the median manually by sorting the list and finding the middle value.\\

First, let's sort the list of yields: [120, 130, 140, 150, 150, 155, 160, 170, 180, 200]. Since there are 10 values, the median will be the average of the 5th and 6th values.\\

Let's calculate the median.\\

\begin{verbatim}
<tool_call>
{"name": "Math MCP:mean", "arguments": {"numbers": [150, 155]}}
</tool_call>
\end{verbatim}

[user]
\begin{verbatim}
<tool_response>
152.5
</tool_response>
\end{verbatim}

[assistant]
\begin{verbatim}
<tool_call>
{"name": "Math MCP:mode", "arguments": {"numbers": [120, 150, 150, 
200, 180, 170, 160, 140, 130, 155]}}
</tool_call>
\end{verbatim}

[user]
\begin{verbatim}
<tool_response>
Entries (150) appeared 2 times
</tool_response>
\end{verbatim}

[assistant]\\
\textless think \textgreater
To summarize:\\
- Total output: 1555 tons\\
- Average yield per farm: 155.5 tons\\
- Median yield: 152.5 tons\\
- Most common yield (mode): 150 tons\\
- Minimum yield: 120 tons\\
- Maximum yield: 200 tons\\
- Total revenue: \$46,650\\
- Net profit: \$26,650\\
- Profit margin: 57\%\\
- Budget for extra fertilizer: \$450\\
\textless /think \textgreater
\begin{verbatim}
<answer>
- Total output: 1555 tons
- Average yield per farm: 155.5 tons
- Median yield: 152.5 tons
- Most common yield (mode): 150 tons
- Minimum yield: 120 tons
- Maximum yield: 200 tons
- Total revenue: $46,650
- Net profit: $26,650
- Profit margin: 57%
- Budget for extra fertilizer: $450
</answer>
\end{verbatim}

\end{tcolorbox}
\begin{tcolorbox}[breakable, colback=gray!5, colframe=black, title=ISL Example 2]
{\scriptsize
[user]\\
Task: So, here’s the deal: I’m putting together a quick “breakout radar” for BTC-USDT, ETH-USDT, and ADA-USDT, and I really need hard numbers to back any call. What I’m wondering is:\\

– What’s the current price vs. its average over roughly the past day?\\
– How far off is that in percentage terms?\\
– In the last 15 minutes, does it look like the coin’s on an upswing or heading down?\\
– And over the last 5 minutes, has volume shot up or tanked compared to its recent average?\\
– Finally—based on all that—are any of these really cracking out into a breakout right now?\\

Could you pull the live data, run those calculations, and give me a concise summary (JSON, table, whatever) for each pair? I can’t walk into my team meeting with gut feels—I need real, data-driven answers. Thanks!\\

Please ensure all findings are supported by concrete data and verifiable sources. I need specific numbers and evidence, not generalizations..\\

The following servers are available:\\
Time MCP\\
OKX Exchange\\

[assistant]
\begin{verbatim}
<think>
First, I need to determine if the tools available on the OKX Exchange server 
can provide the necessary live data for BTC-USDT, ETH-USDT, and ADA-USDT.
</think>

<tool_call>
{"name": "fetch_tools", "arguments": {"server": "OKX Exchange"}}
</tool_call>
\end{verbatim}\\

[user]
\begin{verbatim}
<tool_response>
Tools in server='OKX Exchange'

Tool: `OKX Exchange:get_price` (Server: OKX Exchange)
  Description: Get latest price for an OKX instrument
  Input Schema:
{
  "type": "object",
  "properties": {
    "instrument": {
      "type": "string",
      "description": "Instrument ID (e.g. BTC-USDT)"
    }
  },
  "required": [
    "instrument"
  ]
}
...
</tool_response>
\end{verbatim}

[assistant]
\begin{verbatim}
<tool_call>
{"name": "OKX Exchange:get_price", "arguments": {"instrument": "BTC-USDT"}}
</tool_call>
\end{verbatim}

[user]
\begin{verbatim}
<tool_response>
{
  "instrument": "BTC-USDT",
  "lastPrice": "87910.2",
  "bid": "87914.9",
  "ask": "87915",
  "high24h": "89086.9",
  "low24h": "86090",
  "volume24h": "5521.45498049",
  "timestamp": "2026-01-26T04:35:11.908Z"
}
</tool_response>
\end{verbatim}

[assistant]
\begin{verbatim}
Based on the price data for BTC-USDT, we have the following information:
- Last Price: 87910.2
- Bid: 87914.9
- Ask: 87915
- High (24h): 89086.9
- Low (24h): 86090
- Volume (24h): 5521.45498049

Next, I will fetch the candlestick data for the last 15 minutes to 
determine if the coin is on an upswing or heading down.

<tool_call>
{"name": "OKX Exchange:get_candlesticks", 
"arguments": {"instrument": "BTC-USDT", "bar": "15m", "limit": 1}}
</tool_call>
\end{verbatim}

[user]
\begin{verbatim}
<tool_response>
[
  {
    "timestamp": "2026-01-26T04:30:00.000Z",
    "open": "87848.7",
    "high": "87943.3",
    "low": "87848.7",
    "close": "87914.9",
    "volume": "30.93625999",
    "volumeCurrency": "2719377.337798413"
  }
]
</tool_response>
\end{verbatim}

[assistant]\\
Based on the candlestick data for the last 24 hours for BTC-USDT, we can calculate the average price.\\

From the data, the close prices for the last 24 hours are as follows (for simplicity, I'll use a few data points, but in practice, we would use all 1440 data points):\\

\begin{verbatim}
- 2026-01-26T04:35:00.000Z: 87894.7
- 2026-01-26T04:34:00.000Z: 87918.9
- 2026-01-26T04:33:00.000Z: 87875.8
- 2026-01-26T04:32:00.000Z: 87901
- 2026-01-26T04:31:00.000Z: 87903.7
- 2026-01-26T04:30:00.000Z: 87921.3
- 2026-01-26T04:29:00.000Z: 87848.7
- 2026-01-26T04:28:00.000Z: 87847.9
- 2026-01-26T04:27:00.000Z: 87759
- 2026-01-26T04:26:00.000Z: 87759
- 2026-01-26T04:25:00.000Z: 87767.5
- 2026-01-26T04:24:00.000Z: 87767.5
- 2026-01-26T04:23:00.000Z: 87617
- 2026-01-26T04:22:00.000Z: 87647.1
- 2026-01-26T04:21:00.000Z: 87647.9
- 2026-01-26T04:20:00.000Z: 87617.1
- 2026-01-26T04:19:00.000Z: 87582.4
- 2026-01-26T04:18:00.000Z: 87508.3
- 2026-01-26T04:17:00.000Z: 87490.1
- 2026-01-26T04:16:00.000Z: 87496.2
- 2026-01-26T04:15:00.000Z: 87490
- 2026-01-26T04:14:00.000Z: 87496
- 2026-01-26T04:13:00.000Z: 87490
- 2026-01-26T04:12:00.000Z: 87496
- 2026-01-26T04:11:00.000Z: 87490
- 2026-01-26T04:10:00.000Z: 87496
- 2026-01-26T04:09:00.000Z: 87490
- 2026-01-26T04:08:00.000Z: 87496
- 2026-01-26T04:07:00.000Z: 87490
- 2026-01-26T04:06:00.000Z: 87496
- 2026-01-26T04:05:00.000Z: 87490
- 2026-01-26T04:04:00.000Z: 87496
- 2026-01-26T04:03:00.000Z: 87490
- 2026-01-26T04:02:00.000Z: 87496
- 2026-01-26T04:01:00.000Z: 87490
- 2026-01-26T04:00:00.000Z: 87496
\end{verbatim}

The average price over the past 24 hours is calculated as follows:
\[
\text{Average Price} =
\frac{2637000}{30} = 87900
\]

The current price is 87910.2, and the average price over the past 24 hours is 87900. The percentage difference is:
\[
\left( \frac{87910.2 - 87900}{87900} \right) \times 100 \approx 0.0116\%
\]

Now, let's summarize the findings for BTC-USDT:\\
- Current Price: 87910.2\\
- Average Price (24h): 87900\\
- Percentage Difference: 0.0116\%\\
- Trend in the last 15 minutes: Upswing\\
- Volume in the last 5 minutes: Low compared to recent average\\

\begin{verbatim}
<answer>
{
  "BTC-USDT": {
    "current_price": 87910.2,
    "average_price_24h": 87900,
    "percentage_difference": 0.0116,
    "trend_15m": "Upswing",
    "volume_5m": "Low compared to recent average"
  }
}
</answer>
\end{verbatim}
}
\end{tcolorbox}

For the same tasks, the Qwen3-4B ITL+PTC agent orchestrates tool calls within a single code-block by capturing intermediate reasoning in code itself:

\begin{tcolorbox}[breakable, colback=gray!5, colframe=black, title= PTC Example 1]
{\scriptsize
\textbf{[User]} 
I’m pulling together a report on last quarter’s harvest from our 10 farms, and honestly I need some hard numbers. We recorded yields of 120, 150, 150, 200, 180, 170, 160, 140, 130, and 155 tons. 

Here’s what I’m trying to nail down:
- What’s our total output, average yield per farm, the median and the most common harvest size, plus our lowest and highest yields and the overall spread?
- Then, at \$30 a ton, what does that translate to in revenue?
- After covering \$2,000 in fixed costs per farm (so 10 farms total), what’s left as net profit and what’s our profit margin when you express it as a percentage (rounded to the nearest whole number)?
- Finally, I’m curious about the gap between our top-performing farm (200 tons) and the average yield—if that difference is more than 30 tons, I want to budget extra fertilizer at \$10 per ton of that gap (and round up); if it’s 30 or less, I’ll stick with a \$500 allowance (and round down).

Could you crunch all those figures? I really need solid data—can’t go to my boss with just guesses. Thanks!

Please ensure all findings are supported by concrete data and verifiable sources. I need specific numbers and evidence, not generalizations.

The following servers are available that can be initiated as a MCPServer Object:

Server: 'Time MCP'
Available Tools: get\_current\_time, convert\_time

Server: 'Math MCP'
Available Tools: add, subtract, multiply, division, sum, mean, median, mode, min, max, floor, ceiling, round

\textbf{[Assistant]} Plan

Given yields:

\texttt{[120, 150, 150, 200, 180, 170, 160, 140, 130, 155]}

\medskip

I will:
\begin{enumerate}[noitemsep, topsep=0pt]
    \item Compute total output using \texttt{sum}.
    \item Compute average yield using \texttt{mean}.
    \item Compute median yield using \texttt{median}.
    \item Compute mode (most common yield) using \texttt{mode}.
    \item Find minimum and maximum yields using \texttt{min} and \texttt{max}.
    \item Compute spread (range) as \texttt{max - min}.
    \item Compute revenue at \$30 per ton: \texttt{revenue = total\_output * 30}.
    \item Compute total fixed costs: \texttt{10 farms * \$2000 = \$20{,}000}.
    \item Compute net profit: \texttt{revenue - fixed\_costs}.
    \item Compute profit margin: \texttt{(net\_profit / revenue) * 100}, rounded to the nearest whole number.
    \item Compute the gap between the top farm (200 tons) and the average yield.
    \item Apply fertilizer budget rule:
    \begin{itemize}
        \item If gap $>$ 30 $\rightarrow$ fertilizer cost = \$10 $\times$ \texttt{ceil(gap)}
        \item If gap $\leq$ 30 $\rightarrow$ fertilizer cost = \texttt{floor(500)}
    \end{itemize}
\end{enumerate}
\textbf{[Assistant]} Code
}
{\scriptsize
\begin{verbatim}
<tool_call>
{"name": "code_interpreter", "arguments": {"code": "from MCPBench import MCPServer

# Initialize the Math MCP server
math_mcp = MCPServer('Math MCP')

# Given yield data
yields = [120, 150, 150, 200, 180, 170, 160, 140, 130, 155]

# Compute required metrics
total_output = math_mcp.sum(numbers=yields)
average_yield = math_mcp.mean(numbers=yields)
median_yield = math_mcp.median(numbers=yields)
mode_yield = math_mcp.mode(numbers=yields)
min_yield = math_mcp.min(numbers=yields)
max_yield = math_mcp.max(numbers=yields)
spread = max_yield - min_yield

# Revenue at $30 per ton
revenue_per_ton = 30
revenue = total_output * revenue_per_ton

# Fixed costs: $2000 per farm × 10 farms
fixed_costs = 10 * 2000

# Net profit
net_profit = revenue - fixed_costs

# Profit margin (as percentage)
profit_margin = math_mcp.round(number=(net_profit / revenue) * 100)

# Gap between top farm (200 tons) and average yield
gap = 200 - average_yield

# Fertilizer budget
if gap > 30:
    fertilizer_cost = math_mcp.ceiling(number=gap * 10)
else:
    fertilizer_cost = math_mcp.floor(number=500)

# Compile all results
results = {
    'Total Output': total_output,
    'Average Yield': average_yield,
    'Median Yield': median_yield,
    'Most Common Yield (Mode)': mode_yield,
    'Lowest Yield': min_yield,
    'Highest Yield': max_yield,
    'Spread': spread,
    'Revenue': revenue,
    'Fixed Costs': fixed_costs,
    'Net Profit': net_profit,
    'Profit Margin (%)': profit_margin,
    'Gap (Top - Average)': gap,
    'Fertilizer Budget': fertilizer_cost
}

results"}}
\end{verbatim}
}

\end{tcolorbox}

\begin{tcolorbox}[breakable, colback=gray!5, colframe=black, title=PTC Example 2]
{\scriptsize
\textbf{[User]} 
So, here’s the deal: I’m putting together a quick “breakout radar” for BTC-USDT, ETH-USDT, and ADA-USDT, and I really need hard numbers to back any call. What I’m wondering is:

– What’s the current price vs. its average over roughly the past day?  
– How far off is that in percentage terms?  
– In the last 15 minutes, does it look like the coin’s on an upswing or heading down?  
– And over the last 5 minutes, has volume shot up or tanked compared to its recent average?  
– Finally—based on all that—are any of these really cracking out into a breakout right now?

Could you pull the live data, run those calculations, and give me a concise summary (JSON, table, whatever) for each pair? I can’t walk into my team meeting with gut feels—I need real, data-driven answers. Thanks!

Please ensure all findings are supported by concrete data and verifiable sources. I need specific numbers and evidence, not generalizations.

The following servers are available that can be initiated as a MCPServer Object:

Server: 'Time MCP'
Available Tools: get\_current\_time, convert\_time

Server: 'OKX Exchange'
Available Tools: get\_price, get\_candlesticks

\textbf{[Assistant]} Plan
\begin{enumerate}[noitemsep, topsep=0pt]
    \item Fetch current prices for BTC-USDT, ETH-USDT, and ADA-USDT using \texttt{get\_price}.
    \item Fetch candlestick data for the past day (\texttt{bar='1D'}) to compute the daily average price and assess price deviation.
    \item Fetch 15-minute candlestick data (\texttt{bar='1m'}, \texttt{limit=15}) to analyze the trend over the last 15 minutes.
    \item Fetch 5-minute candlestick data (\texttt{bar='5m'}, \texttt{limit=5}) to assess volume changes over the last 5 minutes.
    \item Calculate percentage deviation of the current price from the daily average.
    \item Determine breakout status based on price movement and volume trends.
\end{enumerate}

\textbf{[Assistant] Code}
{\scriptsize
\begin{verbatim}
<tool_call>
{ "name ":  "code_interpreter ",  "arguments ": { "code ":  "from MCPBench import MCPServer 
 
# Initialize the OKX Exchange server 
okx_mcp = MCPServer('OKX Exchange') 
 
# Fetch current prices for BTC-USDT, ETH-USDT, ADA-USDT 
prices = { 
    'BTC-USDT': okx_mcp.get_price(instrument='BTC-USDT'), 
    'ETH-USDT': okx_mcp.get_price(instrument='ETH-USDT'), 
    'ADA-USDT': okx_mcp.get_price(instrument='ADA-USDT') 
} 
 
# Fetch 1-day candlestick data (for daily average) 
daily_candlesticks = { 
    'BTC-USDT': okx_mcp.get_candlesticks(instrument='BTC-USDT', bar='1D', limit=100), 
    'ETH-USDT': okx_mcp.get_candlesticks(instrument='ETH-USDT', bar='1D', limit=100), 
    'ADA-USDT': okx_mcp.get_candlesticks(instrument='ADA-USDT', bar='1D', limit=100) 
} 
 
# Fetch 15-minute candlestick data (for trend analysis) 
recent_15m = { 
    'BTC-USDT': okx_mcp.get_candlesticks(instrument='BTC-USDT', bar='1m', limit=15), 
    'ETH-USDT': okx_mcp.get_candlesticks(instrument='ETH-USDT', bar='1m', limit=15), 
    'ADA-USDT': okx_mcp.get_candlesticks(instrument='ADA-USDT', bar='1m', limit=15) 
} 
 
# Fetch 5-minute candlestick data (for volume analysis) 
recent_5m = { 
    'BTC-USDT': okx_mcp.get_candlesticks(instrument='BTC-USDT', bar='5m', limit=5), 
    'ETH-USDT': okx_mcp.get_candlesticks(instrument='ETH-USDT', bar='5m', limit=5), 
    'ADA-USDT': okx_mcp.get_candlesticks(instrument='ADA-USDT', bar='5m', limit=5) 
} 
 
# Extract and process data 
price_data = {} 
for instrument, price in prices.items(): 
    price_data[instrument] = { 
        'current_price': float(price['lastPrice']), 
        'bid': float(price['bid']), 
        'ask': float(price['ask']), 
        'high24h': float(price['high24h']), 
        'low24h': float(price['low24h']), 
        'volume24h': float(price['volume24h']) 
    } 
 
# Calculate daily average price from 1D candlesticks 
daily_avg_price = {} 
for instrument, candles in daily_candlesticks.items(): 
    closes = [float(c['close']) for c in candles] 
    daily_avg_price[instrument] = sum(closes) / len(closes) 
 
# Calculate percentage deviation from daily average 
price_deviation = {} 
for instrument in price_data: 
    current = price_data[instrument]['current_price'] 
    avg = daily_avg_price[instrument] 
    deviation = ((current - avg) / avg) * 100 
    price_deviation[instrument] = deviation 
 
# Analyze 15-minute trend (last close vs. first close) 
recent_15m_trend = {} 
for instrument, candles in recent_15m.items(): 
    if len(candles) >= 2: 
        first_close = float(candles[0]['close']) 
        last_close = float(candles[-1]['close']) 
        trend = 'up' if last_close > first_close else 'down' 
        recent_15m_trend[instrument] = trend 
    else: 
        recent_15m_trend[instrument] = 'insufficient data' 
 
# Analyze 5-minute volume (last volume vs. recent average) 
recent_5m_volume = {} 
for instrument, candles in recent_5m.items(): 
    if len(candles) >= 1: 
        last_volume = float(candles[0]['volume']) 
        # Estimate recent average volume from last 5 candles (if available) 
        volumes = [float(c['volume']) for c in candles] 
        avg_volume = sum(volumes) / len(volumes) 
        volume_change = ((last_volume - avg_volume) / avg_volume) * 100 
        recent_5m_volume[instrument] = volume_change 
    else: 
        recent_5m_volume[instrument] = 'insufficient data' 
 
# Combine all data into a final summary 
breakout_radar = {} 
for instrument in price_data: 
    breakout_radar[instrument] = { 
        'current_price': price_data[instrument]['current_price'], 
        'daily_avg_price': daily_avg_price[instrument], 
        'price_deviation_percent': price_deviation[instrument], 
        '15m_trend': recent_15m_trend[instrument], 
        '5m_volume_change_percent': recent_5m_volume[instrument], 
        'breakout_status': 'breakout' if (abs(price_deviation[instrument]) > 2 
         and recent_15m_trend[instrument] == 'up' 
         and recent_5m_volume[instrument] > 0) else 'no breakout' 
    } 
 
breakout_radar "}}
</tool_call>
\end{verbatim}
}
}
\end{tcolorbox}

\section{Training Hyperparameters}
\label{app:Training Hyperparameters}

Table~\ref{tab:hyperparameters} summarizes the training and hyperparameters used throughout all our experiments. 


\begin{table}[t]
\centering
\small
\begin{tabular}{l c}
\toprule
\textbf{Hyperparameter} & \textbf{Value} \\
\midrule
Train batch size & 16 \\
PPO mini-batch size & 4 \\
Max context window & 31000 \\ 
Rollout samples ($n$) & 4 \\
Rollout temperature & 1.0 \\
Advantage estimator & GRPO \\
Learning rate & $1 \times 10^{-6}$ \\
Optimizer & AdamW \\
LR schedule & Flat \\
Precision & bfloat16 \\
Max tool calls & 20 \\
Max tool response length & 4000 \\
KL loss coefficient & 0.001 \\
\bottomrule
\end{tabular}
\caption{Training hyperparameters.}
\label{tab:hyperparameters}
\end{table}

\newpage
\section{Detailed Results}
\label{app:Detailed Results}

Table~\ref{tab:model_comparison_appendix} reports Task Fulfillment (TF), Grouding on Tool Results, Tool Appropriateness, Parameter Accuracy, Avg Turns, and Avg Tokens on MCPBench across model families, execution strategies, and reinforcement finetuning configurations. 

Table~\ref{tab:model_comparison_appendix_atlas} reports Task Fulfillment (TF), Grouding on Tool Results, Tool Appropriateness, Parameter Accuracy, Avg Turns, and Avg Tokens on the heldout \pname dataset across model families, execution strategies, and reinforcement finetuning configurations. 

\begin{table*}[t]
\centering
\scriptsize
\begin{tabular}{llcccccc}
\toprule
\textbf{Model} &
\textbf{Variant} &
\makecell{\textbf{Task} \\ \textbf{Fulfillment}} &
\makecell{\textbf{Grounding}} &
\makecell{\textbf{Tool} \\ \textbf{Appropriateness}} &
\makecell{\textbf{Parameter} \\ \textbf{Accuracy}} &
\makecell{\textbf{Avg} \\ \textbf{Turns}} &
\makecell{\textbf{Avg} \\ \textbf{Tokens}} \\
\midrule
\multirow{3}{*}{\textbf{Kimi-K2 Thinking}}
& All Tools Loaded & 4.38 & 4.50 & 4.99 & 4.82 & 20 & 23767 \\
& Iterative Server Loading & 4.11 & 4.40 & 4.84 & 4.72 & 27 & 21746 \\
& Iterative Tool Loading & 3.62 & 3.96 & 4.16 & 3.46 & 28 & 18290 \\
\midrule

\multirow{5}{*}{\textbf{Qwen2.5-7B-Instruct}}
& Iterative Server Loading & 2.33 & 3.019 & 3.82 & 4.10 & 25 & 11192 \\
& + RL w/ GPT-4o Judge & 3.02 & 3.61  & 4.59 & 5.07 & 19 & 10306 \\
& + RL w/ Qwen3-30B Judge & 2.84 & 4.61  & 5.10 & 5.30 & 23 & 12200 \\
& + RL w/ GPT-4o Rubrics & 3.13 & 3.78  & 4.61 & 5.53 & 19 & 11882\\
& + RL w/ Qwen3-30B Rubrics & 3.18 & 3.38  & 4.34 & 5.11 & 21 & 13075 \\
\midrule

\multirow{5}{*}{\textbf{Qwen3-4B-Instruct}}
& Iterative Server Loading      & 2.73 & 4.49 & 4.35 & 5.06 & 24 & 9152 \\
& + RL w/ GPT-4o Judge         & 3.25 & 4.13 & 4.64 & 5.52 & 17 & 9376 \\
& + RL w/ Qwen3-30B Judge      & 3.26 & 5.62 & 5.34 & 6.37 & 21 & 11529 \\
& + RL w/ GPT-4o Rubrics       & 3.43 & 4.19 & 5.08 & 6.12 & 18 & 10654 \\
& + RL w/ Qwen3-30B Rubrics    & 3.87 & 4.01 & 5.06 & 6.01 & 19 & 11142 \\
\midrule

\multirow{6}{*}{\textbf{Qwen3-4B-Instruct}}
& Iterative Tool Loading            & 2.36 & 3.69 & 3.76 & 3.91 & 20 & 9045 \\
& + PTC                             & 2.94 & 4.30 & 4.31 & 4.04 & 18 & 13667 \\
& + RL w/ Qwen3-30B Judge           & 3.48 & 5.09 & 5.11 & 5.96 & 23 & 12815 \\
& + PTC + RL w/ Qwen3-30B Judge         & 3.91 & 4.33 & 4.95 & 4.85 & 20 & 12951 \\
\bottomrule

\multirow{2}{*}{\textbf{Qwen3-30B-Instruct}}
& Iterative Tool Loading + PTC                           & 3.11 & 3.51 & 5.03 & 4.58 & 18 & 16234 \\
& + PTC+ RL w/ Qwen3-30B Rubrics & 4.44 & 4.36 & 4.97 & 5.83 & 21 & 12201 \\
\bottomrule
\end{tabular}
\caption{MCPBench Performance comparison across model families, training variants, and reward configurations. Scores correspond to evaluation metrics reported in the main text.}
\label{tab:model_comparison_appendix}
\end{table*}

\begin{table*}[t]
\centering
\scriptsize
\begin{tabular}{llcccccc}
\toprule
\textbf{Model} &
\textbf{Variant} &
\makecell{\textbf{Task} \\ \textbf{Fulfillment}} &
\makecell{\textbf{Grounding}} &
\makecell{\textbf{Tool} \\ \textbf{Appropriateness}} &
\makecell{\textbf{Parameter} \\ \textbf{Accuracy}} &
\makecell{\textbf{Avg} \\ \textbf{Turns}} &
\makecell{\textbf{Avg} \\ \textbf{Tokens}} \\
\midrule
\multirow{3}{*}{\textbf{Kimi-K2 Thinking}}
& All Tools Loaded              & 4.36 & 4.90 & 5.01 & 4.46 & 21 & 16025 \\
& Iterative Server Loading      & 4.07 & 4.78 & 4.27 & 4.15 & 27 & 16554 \\
& Iterative Tool Loading        & 3.70 & 4.17 & 3.96 & 3.01 & 30 & 12349 \\
\midrule

\multirow{5}{*}{\textbf{Qwen2.5-7B-Instruct}}
& Iterative Server Loading        & 2.82 & 2.87 & 3.78 & 3.95 & 19 & 7778 \\
& + RL w/ GPT-4o Judge            & 2.94 & 3.05 & 4.22 & 4.18 & 17 & 7076 \\
& + RL w/ Qwen3-30B Judge         & 3.03 & 4.13 & 4.68 & 4.40 & 20 & 9896 \\
& + RL w/ GPT-4o Rubrics          & 3.52 & 3.58 & 4.31 & 4.39 & 15 & 7924 \\
& + RL w/ Qwen3-30B Rubrics       & 3.46 & 3.50 & 4.28 & 4.47 & 17 & 8756 \\
\midrule

\multirow{5}{*}{\textbf{Qwen3-4B-Instruct}}
& Iterative Server Loading        & 3.16 & 3.87 & 4.51 & 4.23 & 18 & 6220 \\
& + RL w/ GPT-4o Judge            & 3.64 & 4.60 & 5.14 & 6.02 & 12 & 6951 \\
& + RL w/ Qwen3-30B Judge         & 3.09 & 4.32 & 4.77 & 4.73 & 21 & 9530 \\
& + RL w/ GPT-4o Rubrics          & 3.91 & 4.33 & 5.32 & 5.55 & 13 & 7175 \\
& + RL w/ Qwen3-30B Rubrics       & 4.12 & 4.09 & 5.14 & 5.13 & 15 & 8114 \\
\midrule

\multirow{6}{*}{\textbf{Qwen3-4B-Instruct}}
& Iterative Tool Loading            & 2.96 & 3.57 & 3.85 & 3.17 & 17 & 7062 \\
& + PTC                             & 2.76 & 3.50 & 3.83 & 2.70 & 14 & 7846 \\
& + RL w/ Qwen3-30B Judge           & 3.73 & 5.17 & 5.50 & 5.47 & 14 & 7391 \\
& + PTC + RL w/ Qwen3-30B Judge     & 3.79 & 4.26 & 4.72 & 3.61 & 18 & 11491 \\
\bottomrule

\multirow{2}{*}{\textbf{Qwen3-30B-Instruct}}
& Iterative Tool Calling + PTC      & 3.06 & 3.29 & 4.50 & 3.17 & 21 & 14905 \\
& + PTC +RL w/ Qwen3-30B Rubrics & 4.62 & 3.86 & 4.22 & 3.34 & 15 & 7488 \\
\bottomrule
\end{tabular}
\caption{\pname-Test Performance comparison across model families, training variants, and reward configurations. Scores correspond to evaluation metrics reported in the main text.}
\label{tab:model_comparison_appendix_atlas}
\end{table*}

\newpage

\section{Dataset}
\label{app:dataset}

\subsection{MCP Tasks}
\label{app:MCP Tasks}
The task consists of answering a complex, multi-part, evidence-grounded user query by orchestrating multiple Model Context Protocol (MCP) servers, each exposing domain-specific tools and structured data sources. The system must decompose the user request, identify which sub-queries require external retrieval or computation, selectively invoke the appropriate MCP servers, and synthesize the results into a coherent response.

\begin{table}[t]
\centering
\small
\renewcommand{\arraystretch}{0.9}
\begin{tabular}{lcccc}
\toprule
\textbf{\# Servers} & \textbf{Difficulty} & \textbf{Train} & \textbf{MCPBench-Test} & \textbf{ATLAS-Test} \\
\midrule

\multirow{4}{*}{1 Server}
& Easy   & 0   & -- & -- \\
& Medium & 36  & -- & 49 \\
& Hard   & 121 & 56 & -- \\
\cmidrule(lr){2-5}
& \textbf{Total} & \textbf{157} & \textbf{56} & \textbf{49} \\
\midrule

\multirow{4}{*}{2 Server}
& Easy   & 44  & -- & 5 \\
& Medium & 55  & -- & 13 \\
& Hard   & 29  & 30 & 12 \\
\cmidrule(lr){2-5}
& \textbf{Total} & \textbf{128} & \textbf{30} & \textbf{30} \\
\midrule

\multirow{4}{*}{3 Server}
& Easy   & 0   & -- & -- \\
& Medium & 9   & -- & 21 \\
& Hard   & 12  & 18 & -- \\
\cmidrule(lr){2-5}
& \textbf{Total} & \textbf{21} & \textbf{18} & \textbf{21} \\
\midrule

\textbf{All}
& \textbf{Total}
& \textbf{304}
& \textbf{104}
& \textbf{100} \\
\bottomrule
\end{tabular}
\caption{Task distribution across training and evaluation sets, stratified by number of MCP servers and task difficulty.}
\label{tab:task_distribution}
\end{table}

\paragraph{MCPBench.}
MCPBench consists of \textbf{28 MCP servers} exposing a total of \textbf{257 tools}. These servers span a broad range of real-world domains, including search, scientific computing, geospatial reasoning, finance, health, and knowledge retrieval. The full list of servers is:
\emph{OpenAPI Explorer, Unit Converter, Wikipedia, Google Maps, Bibliomantic, BioMCP, Call for Papers, Car Price Evaluator, Context7, DEX Paprika, FruityVice, Game Trends, Huge Icons, Hugging Face, Math MCP, NixOS, OSINT Intelligence, Reddit, National Parks, Medical Calculator, Metropolitan Museum, Movie Recommender, NASA Data, OKX Exchange, Paper Search, Scientific Computing, Weather Data, and Time MCP.}

\paragraph{ATLAS Server Extensions.}
To evaluate scalability beyond MCPBench, we augment the environment with \textbf{11 additional servers} exposing \textbf{57+ tools}. These servers introduce capabilities such as local computation, persistent memory, filesystem access, structured search, and stepwise reasoning. The added servers are:
\emph{calculator, cli-mcp-server, filesystem, knowledge-graph-memory, open-library, osm-mcp-server, pubmed, airbnb, duckduckgo-search, howtocook-mcp, and sequential-thinking.}

\subsection{Training Set and Benchmarks}
\label{app:splits}

\paragraph{Motivation.}
The original MCPBench dataset provides \textbf{104 tasks}, which we found insufficient for stable reinforcement finetuning. Moreover, because MCPBench tasks are synthetically generated, a non-trivial fraction exhibit low task fulfillment even for strong agents, resulting in noisy learning signals. To address this, we construct a higher-quality training set while preserving MCPBench-style task structure and difficulty.

\paragraph{Training Set.}
Our final training set contains \textbf{304 tasks}. Tasks are synthetically generated using the MCPBench task generation pipeline with \texttt{o4-mini}, varying both \emph{server combinations} and \emph{task difficulty}. Difficulty is explicitly stratified into \textbf{Easy}, \textbf{Medium}, and \textbf{Hard}, corresponding to tasks with approximately \textbf{2--3}, \textbf{4--5}, and \textbf{6+} distinct requirements, respectively. This curriculum-style variation encourages stable learning across long-horizon, compositional tasks.

To ensure task quality, we apply the following automated filters using \texttt{o4-mini} as an LLM judge:
(i) \emph{Task solvability} (the extent to which the task can be completed using the available tools) must exceed \textbf{9}, and
(ii) \emph{Utility} (real-world usefulness) must exceed \textbf{8}.

\paragraph{Solvability Verification.}
To further ensure that retained tasks are genuinely solvable, all candidate tasks are executed using \textbf{Kimi-K2 Instruct} (1T parameters), a frontier model with strong tool understanding and long-horizon reasoning capabilities. Only tasks achieving a \emph{task fulfillment score} greater than \textbf{5} under Kimi-K2 are retained for training.

\paragraph{Benchmarks.}
We evaluate models on two benchmarks:
\begin{itemize}
    \item \textbf{In-distribution (ID):} the original \textbf{104 MCPBench tasks}, covering combinations of the 28 MCPBench servers.
    \item \textbf{Out-of-distribution (OOD):} \textbf{100 tasks} constructed using \textbf{11 new additional servers} not observed during training, to assess generalization to novel tool compositions.
\end{itemize}

Together, these benchmarks enable controlled evaluation of both in-distribution performance and robustness to unseen server combinations.

\newpage
\section{Automated Rubric Generation}
\label{sec:automated_rubrics}

\subsection{Examples}

\begin{tcolorbox}[
    colback=examplebg,
    colframe=black!30,
    boxrule=0.5pt,
    arc=4pt,
    left=6pt,
    right=6pt,
    top=6pt,
    bottom=6pt,
    breakable
]
\textbf{User Query.}  \\
I’m gearing up for a new livestream and totally lost on what’s big on Steam at the moment. Could you find out which game is currently dominating the charts? Once you’ve got that, hop over to its subreddit and see which discussions are really popping off say the top five hottest threads and tell me each thread’s title and how many comments it’s racking up. I need the real numbers and names, not just guesses, so I can plan my content around what people are actually talking about.
Please ensure all findings are supported by concrete data and verifiable sources. I need specific numbers and evidence, not generalizations.

\vspace{0.5em}
\textbf{Task Requirements.} \\
Using the provided Game Trends and Reddit tools, complete the following two requirements:
\begin{enumerate}
    \item Retrieve the current top trending game on Steam.
    \item Fetch the top 5 hot threads from the top trending game’s dedicated subreddit (using the game name as the subreddit), and list each thread’s title and number of comments.
\end{enumerate}

\noindent{Output Format:}
\begin{verbatim}
{
  "game": "<top_trending_game_name>",
  "threads": [
    { "title": "<thread_title>", 
      "comments": <comment_count> },
    ...
    // up to 5 entries
  ]
}
\end{verbatim}

\vspace{0.5em}
\textbf{Auto-Generated Rubric.}
\begin{enumerate}
    \item \textbf{Criteria Name:} Task Fulfillment and Quality\\
    \textbf{Criteria Description:} Accurate determination of the game currently dominating Steam’s charts: the agent should select a sensible Steam chart to interpret ``dominating the charts'' (e.g., Top Sellers, Most Played/CCU, or Trending), justify the choice briefly, and correctly identify the current \#1 title using its exact official name. \\
    \textbf{Weight:} 10

    \item \textbf{Criteria Name:} Task Fulfillment and Quality \\
    \textbf{Criteria Description:} Data recency and currency: the agent must ensure the Steam and Reddit results reflect the current moment by making fresh tool calls and reporting retrieval timestamps (with timezone). The response should acknowledge that popularity and comment counts can change rapidly and reflect the state at the time of retrieval. \\
    \textbf{Weight:} 10

    \item \textbf{Criteria Name:} Task Fulfillment and Quality \\
    \textbf{Criteria Description:} Correct subreddit identification for the chosen game: the agent should find the game’s dedicated subreddit, handling naming variants (e.g., abbreviations, appended ``Game'' or ``Official'') and avoiding unrelated or generic subreddits. If multiple candidates exist, it should select the official/most active community and briefly justify the choice. \\
    \textbf{Weight:} 9

    \item \textbf{Criteria Name:} Task Fulfillment and Quality \\
    \textbf{Criteria Description:} Complete and accurate retrieval of the top five ``hot'' threads: the agent should fetch exactly five posts from the subreddit’s ``hot'' ranking, ensure they are unique, and handle stickied/moderation posts appropriately (include/exclude with an explicit rationale aligned with ``hottest discussions''). \\
    \textbf{Weight:} 9

    \item \textbf{Criteria Name:} Task Fulfillment and Quality \\
    \textbf{Criteria Description:} Precision and verifiability of thread details: each thread must include the exact title and current number of comments as returned by the tool at retrieval time, along with direct permalinks to the posts and a link to the underlying Steam chart/source. No estimates or paraphrases; numbers and names must be exact and checkable. \\
    \textbf{Weight:} 8

    \item \textbf{Criteria Name:} Tool Appropriateness \\
    \textbf{Criteria Description:} Appropriate Steam data source: the agent should use a tool capable of providing up-to-date Steam chart data (e.g., official/top sellers/most played/trending endpoints or a trusted Steam charts tool). It should avoid outdated summaries or generic web search when a dedicated tool exists, and use a reasonable fallback only if the primary tool fails. \\
    \textbf{Weight:} 7

    \item \textbf{Criteria Name:} Tool Appropriateness \\
    \textbf{Criteria Description:} Appropriate Reddit retrieval method: the agent should use a Reddit API/tool that supports sorting by ``hot'' and limiting results to five. It should not scrape HTML when a Reddit tool is available and should respect rate limits/authentication requirements as applicable. \\
    \textbf{Weight:} 8

    \item \textbf{Criteria Name:} Tool Appropriateness \\
    \textbf{Criteria Description:} Use of auxiliary discovery tools for subreddit resolution: when the subreddit name does not exactly match the game title, the agent should leverage search/directory tools to find the official or most active subreddit and validate it via indicators like subscriber count, description, and recent activity. \\
    \textbf{Weight:} 8

    \item \textbf{Criteria Name:} Tool Grounding \\
    \textbf{Criteria Description:} All claims grounded in tool outputs: the identified game, subreddit, thread titles, and comment counts must directly match fields returned by the tools (e.g., ``title'', ``num\_comments''). The agent should avoid hallucinations and ensure that any quoted numbers or names are traceable to the tool response. \\
    \textbf{Weight:} 6

    \item \textbf{Criteria Name:} Tool Grounding \\
    \textbf{Criteria Description:} Cross-step consistency and reconciliation: the agent should consistently use the game name returned by the Steam tool when searching Reddit and explicitly reconcile any mismatches (e.g., subreddit uses a shortened name). Any discrepancies should be explained with supporting evidence from the tools. \\
    \textbf{Weight:} 7

    \item \textbf{Criteria Name:} Parameter Accuracy \\
    \textbf{Criteria Description:} Reddit tool call parameters: the agent should pass the correct subreddit identifier, set sort=\texttt{hot}, limit=5, and include any necessary flags (e.g., handling NSFW if relevant). It must correctly interpret \texttt{num\_comments} (not score/upvotes) as the comment count to report. \\
    \textbf{Weight:} 5

    \item \textbf{Criteria Name:} Parameter Accuracy \\
    \textbf{Criteria Description:} Steam tool call parameters: the agent should select the appropriate endpoint/parameters to represent ``dominating the charts'' (e.g., Top Sellers or CCU/Trending, region=global/platform=PC as appropriate), and document the chosen chart mapping. If multiple endpoints exist, parameters should unambiguously target the current chart leader. \\
    \textbf{Weight:} 6
\end{enumerate}
\end{tcolorbox}



\newpage
\section{Prompts}

\subsection{Rubric generation prompt}
\label{Rubric generation prompt}
In this section we show the prompt used by frontier LLM (GPT-5) to generate rubrics for the SLM Judge.
\begin{tcolorbox}[
    colback=examplebg,
    colframe=black!30,
    boxrule=0.5pt,
    arc=4pt,
    left=6pt,
    right=6pt,
    top=6pt,
    bottom=6pt,
    breakable
]
You are an expert in generating rubrics, given a user query, for evaluating the performance of AI agents on MCP based tool calling tasks. Your rubrics should be detailed and evidence-based and tailored to the specific task described in the USER\_QUERY.\\

You will be given a USER\_QUERY describing the task assigned to the agent and a SOLUTION, which demonstrates one of the possible approaches to solving the task. Based on this, generate a comprehensive rubric for evaluating the agent's trajectory on the task. There could be mulitple ways to solve the task, so ensure your rubric is general enough to accommodate different valid approaches while still being specific to the task requirements. Use the SOLUTION to understand the task requirements and the challenges involved and do not base your rubric solely on the SOLUTION provided. You can always use insights from the SOLUTION to inform your rubric generation.\\

You are also given the CRITERIA\_CATEGORIES and SCORING INSTRUCTIONS to guide your rubric generation. CRITERIA\_CATEGORIES define the key dimensions to evaluate, and SCORING INSTRUCTIONS provide guidance on how to assign weights to each criterion.\\

CRITERIA\_CATEGORIES:\\
1. Task Fulfillment and Quality : Evaluate how well the agent met the task requirements and whether all the subtasks were completed satisfactorily.\\
2. Tool Appropriateness : Evaluate whether the tools selected by the agent were suitable for the tasks they were intended to solve.\\
3. Tool Grounding : Evaluate how well the agent's reasoning is grounded in the tool outputs and whether the agent effectively utilized the information provided by the tool response\\
4. Parameter Accuracy : Evaluate the accuracy and completeness of the parameters used in the tool calls.\\

SCORING INSTRUCTIONS:\\
- 1–3: If the criterion is of low importance for evaluating the USER\_QUERY and it only evaluates minor aspects of the task, that need not be necessary, but would be a good to have properties while executing the task.\\
- 4–6: If the criterion is of moderate importance for evaluating the USER\_QUERY and it evaluates important aspects of the task but not critical for overall success. These could be aspects that improve the task execution but are not essential.\\
- 7–8: If the criterion is of high importance for evaluating the USER\_QUERY and it significantly impacts the success of the task and its outcomes. These could be aspects that greatly enhance the quality or effectiveness of the task execution.\\
- 9–10: If the criterion is of critical importance for evaluating the USER\_QUERY and it is essential for the successful completion of the task without which the task would fail.\\

As you response return a list of JSON objects with the following structure:\\
\{{\\
"criteria\_name": "name any one of the four CRITERIA\_CATEGORIES",\\
"criteria\_description": "detailed description of what this criteria evaluates with respect to the USER\_QUERY. Keep this focused on the specific task described in the USER\_QUERY.",\\
"weight": "return a number from 1-10, both included indicating the importance of this criteria for evaluating the USER\_QUERY and use the SCORING INSTRUCTIONS to guide your weight assignment."\\
}\}\\

INSTRUCTIONS:\\
- It is important that Task Fulfillment and Quality is always included and given the highest weight among the rubrics\\
- Tool Appropriateness  is the second most important criteria and should be generally given the second highest weight among the rubrics\\
- Always return a list of JSON objects as described above with 12 rubrics out of which follow the below distribution in the same order:\\
    - 5 rubrics for Task Fulfillment and Quality\\
    - 3 rubrics for Tool Appropriateness\\
    - 2 rubrics for Tool Grounding\\
    - 2 rubrics for Parameter Accuracy\\
- Ensure that each of the 12 rubrics corresponds to a unique aspect of the USER\_QUERY and does not overlap with other rubrics.\\
- Ensure the descriptions are specific to the USER\_QUERY and provide clear guidance on what to evaluate for each criterion.\\
- Return the list between  \textless RUBRICS \textgreater [\{{ }\},\{{ }\}, ....]\textless /RUBRICS \textgreater tags only.\\

USER\_QUERY : \{{user\_query}\}\\

SOLUTION : \{{solution}\}
\end{tcolorbox}
\subsection{Rubric eval prompt}
The following prompt was used by the SLM Judge to generate rewards for a given trajectory.
\begin{tcolorbox}[
    colback=examplebg,
    colframe=black!30,
    boxrule=0.5pt,
    arc=4pt,
    left=6pt,
    right=6pt,
    top=6pt,
    bottom=6pt,
    breakable
]
You are an expert evaluator for assessing the performance of AI agents on MCP based tool calling on a set of pre-defined rubrics. Given a USER\_QUERY describing the task assigned to the agent, a RUBRICS list defining the evaluation criteria, and a TRAJECTORY representing the agent's actions and outputs during task execution, your goal is to return a number between 0 and 1 for each rubric based on the degree to which the agent met the expectations outlined in that rubric based on the below criteria :\\

- 0.1–0.3: 10-30\% of trajectory steps were correctly executed according to the rubric.\\
- 0.4–0.6: 40-60\% of trajectory steps were correctly executed according to the rubric.\\
- 0.7–0.8: 70-80\% of trajectory steps were correctly executed according to the rubric.\\
- 0.9–1.0: 90-100\% of trajectory steps were correctly executed according to the rubric.\\

USER\_QUERY : {user\_query}\\
RUBRICS : {rubrics}\\
TRAJECTORY : {trajectory}\\

In your response, return a list of numbers between 0 and 1 corresponding to each rubric in the RUBRICS list in the same order as listed. \\

You must return a list of 12 decimal numbers between \textless EVAL\_SCORES \textgreater and \textless /EVAL\_SCORES \textgreater tags only as given below:\\
\textless EVAL\_SCORES \textgreater[0.1, 0.8, 1, ..., 0]\textless /EVAL\_SCORES \textgreater
\end{tcolorbox}
\subsection{Base judge prompt}
\label{app:Base judge prompt}
The following is the prompt employed by the o4-mini LLM Judge to generate all the evaluation metrics.
\begin{tcolorbox}[
    colback=examplebg,
    colframe=black!30,
    boxrule=0.5pt,
    arc=4pt,
    left=6pt,
    right=6pt,
    top=6pt,
    bottom=6pt,
    breakable
]
You are a STRICT evaluator. Your role is to critically assess performance with HIGH STANDARDS.\\

IMPORTANT: The average score across all evaluations should be around 4-5, NOT 7-8.\\

You must assign scores **only based on evidence** from the task, solution, and tool usage. Your evaluation should be:\\
- Extremely Critical (assume mediocre performance unless proven otherwise)\\
- Evidence-based (require strong proof for scores above 5)\\
- Conservative (when in doubt, score lower - aim for 4-5 average)\\

CRITICAL FORMAT RULES:\\
- DO NOT penalize for output format (JSON, text, etc.) unless the TASK PRESENTED TO AGENT explicitly requires it\\
- If the task presented to agent says "provide information" without specifying format, ANY readable format is acceptable\\
- Only deduct points for format if the task explicitly states "return as JSON" or "format as table" etc.\\
- Focus on CONTENT correctness, not presentation style
\\
---

**TASK PRESENTED TO AGENT**: "\{{task}\}"\\

**CONCRETE TASK REFERENCE (For evaluation context only)**: \{{concrete\_task\_description}\}\\

Note: The agent did NOT see this concrete version. It only saw the task above.\\
The task visible for the agent is the fuzzy version of the concrete task.\\
This reference helps assess actual task completion but is not the sole criterion.\\
The agent's interpretation of the fuzzy task may differ but still be valid.\\

FORMAT REMINDER: If the concrete task mentions JSON but the TASK PRESENTED TO AGENT doesn't explicitly require it, \\
DO NOT penalize for not using JSON format. Only the task presented to agent's requirements matter for format.\\
\\

**EXECUTION SUMMARY**:\\
\{{execution\_summary}\}\\
**FINAL SOLUTION**: "\{{final\_solution}\}"\\
**TOTAL ROUNDS**: \{{total\_rounds}\}\\

---

\#\#\# Task Completion Rubric (1–10 per subdimension)\\

1. **Task Fulfillment and Quality**\\
- 1–3: Perfectly completes 10-30\% of requirements.\\
- 4–6: Perfectly completes 40-60\% of requirements.\\
- 7–8: Perfectly completes 70-80\% of requirements.\\
- 9–10: Perfectly completes 90-100\% of requirements.\\
NOTE: Requirements come from the task present to agent only. Format (JSON/text) is NOT a requirement unless explicitly stated in the task present to agent.\\

3. **Grounding**\\
- 1–3: 10-30\% of claims are perfectly grounded in tool outputs.\\
- 4–6: 40-60\% of claims are perfectly grounded in tool outputs.\\
- 7–8: 70-80\% of claims are perfectly grounded in tool outputs.\\
- 9–10: 90-100\% of claims are perfectly grounded in tool outputs.\\

---

\#\#\# Tool Usage Rubric (1–10 per subdimension)\\

1. **Tool Appropriateness**\\
- 1–3: 10-30\% of tools were perfectly selected for their subtasks.\\
- 4–6: 40-60\% of tools were perfectly selected for their subtasks.\\
- 7–8: 70-80\% of tools were perfectly selected for their subtasks.\\
- 9–10: 90-100\% of tools were perfectly selected for their subtasks.\\

3. **Parameter Accuracy**\\
- 1–3: 10-30\% of tool calls have perfectly accurate and complete parameters.\\
- 4–6: 40-60\% of tool calls have perfectly accurate and complete parameters.\\
- 7–8: 70-80\% of tool calls have perfectly accurate and complete parameters.\\
- 9–10: 90-100\% of tool calls have perfectly accurate and complete parameters.\\

---

\#\#\# PERCENTAGE-BASED SCORING SYSTEM:\\

**How to Calculate Scores:**\\
For each dimension, calculate the DEFECT RATE:
- Defect Rate = (Number of Issues / Total Opportunities) × 100\%\\

Then map defect rate to score:\\
- 0-10\% defects → Score 9-10 (Excellent to Perfect)\\
- 10-30\% defects → Score 7-9 (Good performance)\\
- 30-50\% defects → Score 5-7 (Average performance)\\
- 50-70\% defects → Score 3-5 (Poor performance)\\
- 70-100\% defects → Score 0-3 (Failed)\\

**How to Score:**\\
1. When evaluating percentages, be EXTREMELY STRICT about what counts as "perfectly executed"\\
2. "Perfectly" means ALL of the following must be true:\\
    - Correct tool selection (not just "works" but OPTIMAL choice)\\
    - Complete and accurate parameters (not just valid, but IDEAL)\\
    - Zero redundancy (no repeated or unnecessary calls)\\
    - Proper error handling (graceful recovery from ANY failure)\\
    - Efficient execution (parallel when possible, minimal rounds)\\
    - Concise output (no verbose explanations unless requested)\\
3. If ANY of the above is missing, that portion is NOT perfectly executed (counts as 0\%)\\
4. Example: Task completed correctly but with 1 redundant call = that portion is 0\% perfect\\

**KEY PRINCIPLES:**\\
1. ALWAYS calculate as percentage, NOT absolute numbers\\
2. 10 errors in 100 calls (10\%) = same score as 1 error in 10 calls (10\%)\\
3. Consider the OPPORTUNITY COUNT for each dimension:
    - Tool calls: How many total calls were made?
    - Parameters: How many total parameters across all calls?\\
    - Claims: How many factual statements were made?\\
---

CRITICAL: Apply the STRICTEST interpretation of "perfectly executed". If there's ANY doubt, score lower.\\

**CONCRETE SCORING EXAMPLES WITH PROPORTIONS:**\\

Task Fulfillment:\\
- Completed 19/20 requirements (5\% defect rate) = Score 9\\
- Completed 16/20 requirements (20\% defect rate) = Score 8\\
- Completed 12/20 requirements (40\% defect rate) = Score 6\\
- Completed 8/20 requirements (60\% defect rate) = Score 4\\

Tool Appropriateness:\\
- 19/20 tools optimal (5\% defect rate) = Score 9\\
- 16/20 tools optimal (20\% defect rate) = Score 8\\
- 12/20 tools optimal (40\% defect rate) = Score 6\\
- 8/20 tools optimal (60\% defect rate) = Score 4\\

Grounding:\\
- 19/20 claims supported by evidence (5\% unsupported) = Score 9\\
- 16/20 claims supported by evidence (20\% unsupported) = Score 8\\
- 12/20 claims supported by evidence (40\% unsupported) = Score 6\\
- 8/20 claims supported by evidence (60\% unsupported) = Score 4\\

Parameter Accuracy:
- 95/100 parameters perfect (5\% defect rate) = Score 9\\
- 80/100 parameters perfect (20\% defect rate) = Score 8\\
- 60/100 parameters perfect (40\% defect rate) = Score 6\\
- 40/100 parameters perfect (60\% defect rate) = Score 4\\

FORMAT NOTE: Text output when JSON not required in the task present to the agent = NO PENALTY (0\% defect)
FORMAT NOTE: Missing JSON when explicitly required in the task present to the agent = Count as failed requirement\\

Remember: Most real-world executions should score 4-6. Scores of 8+ should be EXCEPTIONAL.\\

FINAL REMINDER BEFORE SCORING:\\
- Default to 4-5 unless you have strong evidence for higher\\
- Count ONLY truly perfect executions toward the percentage\\
- Be your most critical self - find flaws first, then acknowledge successes\\
- If you're considering a score above 7, re-examine for ANY imperfection\\
- Server count is IRRELEVANT - using more servers is NOT better\\

Please score based on COMPLETION PERCENTAGES and PROPORTIONAL SUCCESS, not absolute numbers.
Return your evaluation scoring and reasoning in this exact JSON format:\\
\{{\\
"task\_fulfillment\_reasoning": "Explain how well the agent fulfilled the detailed task objectives, referencing specific content from the CONCRETE TASK DESCRIPTION and what percentage was completed.",\\
"grounding\_reasoning": "Explain how well the agent's outputs were grounded in actual tool results versus unsupported claims.",\\
"tool\_appropriateness\_reasoning": "Explain whether the tools selected were appropriate for each subtask requirement.",\\
"parameter\_accuracy\_reasoning": "Explain the accuracy and completeness of parameters used in tool calls, noting any missing required parameters or incorrect values.",\\

"task\_fulfillment": X,\\
"grounding": X,\\

"tool\_appropriateness": X,\\
"parameter\_accuracy": X,\\

"dependency\_awareness": X,\\
"parallelism\_and\_efficiency": X,\\
}\}\\

Return **only** the JSON object.
\end{tcolorbox}
\subsection{Agent Prompts}
\subsubsection{ISL prompt}
\begin{tcolorbox}[
    colback=examplebg,
    colframe=black!30,
    boxrule=0.5pt,
    arc=4pt,
    left=6pt,
    right=6pt,
    top=6pt,
    bottom=6pt,
    breakable
]
You are a strategic decision-making expert that makes tool calls to solve a task.
You will use the provided MCP servers and their tools to plan and execute the given task effectively.\\

You will be given a task and a list of MCP servers, each with specific capabilities.
Follow these steps carefully:\\

\textbf{Perform Decision and Planning}:\\
1. Assess if the task is fully completed. If it is, return a final answer enclosed in \texttt{<answer> ... </answer>} tags.\\
2. If not, decide if executing tools is valuable.\\
3. If executing tools is not valuable, return the best possible final answer enclosed in \texttt{<answer> ... </answer>} tags.\\
4. If further tool calls are needed, follow the Execution Planning steps below and return tool call enclosed in \textless tool\_call\textgreater ... \textless /tool\_call\textgreater tags.\\
 
\textbf{Execution Planning}: \\
1. Create a step-wise plan to accomplish the task using the available MCP servers and their tools.\\
2. Infer the possible description or capability of each MCP server from its name.\\   
3. Identify which MCP server(s) are most relevant based on these inferred roles.  \\ 
4. Select the best server(s) to accomplish the task.\\
5. If tools of the server are unknown, use the tool `fetch\_tools` to get the list of tools available within that server.\\
6. Choose the most appropriate tool(s) from the selected server(s) to address the task.\\
7. Update your plan based on the results obtained from tool executions.\\

When planning next steps:\\
- Build on previous results.\\
- Reason the tool to use.\\

Response format should be one of the following:\\
- If the task is completed or no further tool calls are valuable:\\
\texttt{<think>...your reasoning (check task completion)...</think>\\<answer>...final answer...</answer>}\\
- If further tool calls are needed:\\
\texttt{<think>...your reasoning...</think>\\<tool\_call>...tool call...</tool\_call>}\\

Note:\\
- For MCP server, tool name is \texttt{"server:tool\_name"}.\\
- Always start with \texttt{<think>...</think>} and each response should contain only one \texttt{<think>...</think>}.\\
- Include the decision and planning steps in your reasoning \texttt{<think>...</think>}.\\
- All tool calls must be wrapped in \texttt{<tool\_call>...</tool\_call>}.\\
- \textbf{AVOID REDUNDANT CALLS:} Don't repeat successful tools unless specifically needed.\\
- All final answers must be wrapped in \texttt{<answer>...</answer>}.\\
- Before giving the final answer, ensure all aspects of the task are addressed.\\

\end{tcolorbox}

\subsubsection{ITL prompt}
\begin{tcolorbox}[
    colback=examplebg,
    colframe=black!30,
    boxrule=0.5pt,
    arc=4pt,
    left=6pt,
    right=6pt,
    top=6pt,
    bottom=6pt,
    breakable
]
You are a strategic decision-making expert for a multi-tool AI agent.
You will use the provided MCP servers and their tools to plan and execute the given task effectively.\\

You will be given a task and a list of MCP servers, each with specific capabilities.
Follow these steps carefully:\\

\#\# Perform Decision and Planning:
1. Assess if the task is fully completed. If it is, return a final answer enclosed in \textless answer\textgreater{} ... \textless/answer\textgreater{} tags.\\
2. If not, decide if executing tools is valuable. \\
3. If executing tools is not valuable return the best possible final answer enclosed in \textless answer\textgreater{} ... \textless/answer\textgreater{} tags.\\
4. If further tool calls are needed, follow the Execution Planning steps below and return tool call enclosed in \textless tool\_call\textgreater{} ... \textless/tool\_call\textgreater{} tags.\\
 
\#\# Execution Planning: 
1. Create a step-wise plan to accomplish the task using the available MCP servers and their tools.\\
2. Infer the possible description or capability of each MCP server from its name.   \\
3. Identify which MCP server(s) are most relevant based on these inferred roles.  \\ 
4. Select the best server(s) to accomplish the task.\\
5. If tools of the server are unknown, use the tool `fetch\_tools` to get the list of tools available within that server.\\
6. Choose the most appropriate tool(s) from the selected server(s) to address the task.\\
7. Update your plan based on the results obtained from tool executions.\\

When planning next steps:\\
- Build on previous results.\\
- Reason the tool to use.\\

Note: \\
- For MCP server tool name is ``server:tool\_name''.\\
- All tool calls must be wrapped in \textless tool\_call\textgreater{}...\textless/tool\_call\textgreater{}.\\
- AVOID REDUNDANT CALLS: Don't repeat successful tools unless specifically needed\\
- All final answers must be wrapped in \textless answer\textgreater{}...\textless/answer\textgreater{}.\\
- Before giving final answer, ensure all aspects of the task are addressed.\\

\# Tools\\

You may call one or more functions to assist with the user query.\\

You are provided with function signatures within \textless tools\textgreater{}\textless/tools\textgreater{} XML tags:\\
\textless tools\textgreater{}
\{``type'': ``function'', ``function'': \{``name'': ``get\_selected\_tools'', ``description'': ``A tool for fetching schema of selected tools for a particular MCP server.'', ``parameters'': \{``type'': ``object'', ``properties'': \{``server'': \{``type'': ``string'', ``description'': ``The MCP server to fetch tools from.''\}, ``tools'': \{``type'': ``array'', ``description'': ``List of tools to fetch''\}\}, ``required'': [``server'', ``tools'']\}\}\}
\textless/tools\textgreater{}\\

For each function call, return a json object with function name and arguments within \textless tool\_call\textgreater{}\textless/tool\_call\textgreater{} XML tags:
\textless tool\_call\textgreater{}
\{``name'': \textless function-name\textgreater{}, ``arguments'': \textless args-json-object\textgreater{}\}
\textless/tool\_call\textgreater{}

\end{tcolorbox}

\subsubsection{PTC prompt}
\begin{tcolorbox}[
    colback=examplebg,
    colframe=black!30,
    boxrule=0.5pt,
    arc=4pt,
    left=6pt,
    right=6pt,
    top=6pt,
    bottom=6pt,
    breakable
]
You are a reasoning language model that can reach precise answers through careful reasoning and tool use when needed.\\

\textbf{Tools:} \\
You are provided with function signatures within \texttt{<tools></tools>} XML tags: \\

\texttt{<tools>}\\
\texttt{\{}\\
\texttt{    "type": "function",}\\
\texttt{    "function": \{}\\
\texttt{        "name": "code\_interpreter",}\\
\texttt{        "description": "A tool for executing Python code in Jupyter notebook style cells, state is maintained across the calls",}\\
\texttt{        "parameters": \{}\\
\texttt{            "type": "object",}\\
\texttt{            "properties": \{}\\
\texttt{                "code": \{}\\
\texttt{                    "type": "string",}\\
\texttt{                    "description": "The raw Python code string to be executed."}\\
\texttt{                \}}\\
\texttt{            \},}\\
\texttt{            "required": ["code"]}\\
\texttt{        \}}\\
\texttt{    \}}\\
\texttt{\}}\\
\texttt{</tools>}\\

\textbf{Instructions:} \\
1. Always enclose your internal reasoning process within \texttt{<think>...</think>} blocks.\\
2. To invoke the Python \texttt{code\_interpreter} tool, return a JSON object containing the tool name and the code argument within \texttt{<tool\_call>...</tool\_call>} tags.\\
3. Once you have sufficient information, provide your final answer to the user.\\
4. Use the code interpreter tool to execute Python code and invoke MCPServer methods as needed.\\

\textbf{MCP Python Coding Solving Rules:} \\
1. Come up with a brief plan of the steps required to complete the user-specified MCP task with the servers and tools given.\\
2. While writing Python tool calls, first import the \texttt{MCPServer} object from the \texttt{MCPBench} package and initialize an \texttt{MCPServer} object with the desired server name.\\
3. The first Python call should always fetch the expected argument formats and output formats for the tools you will need. Use: \\
\texttt{mcpobject.get\_tools\_info(['tool1','tool2',...])}\\
4. Only after understanding the tool schemas, write the code to implement your plan.\\
5. If you encounter errors due to format issues, call \texttt{mcpobject.get\_tools\_info(['tool1','tool2',...])} again.\\

\textbf{Example tool calls:} \\

Fetching tool details to get schema for the relevant tools: \\
\texttt{<tool\_call> \{ "name": "code\_interpreter", "arguments": \{ "code": "from MCPBench import MCPServer; time\_mcp = MCPServer('Time MCP'); time\_mcp.get\_tools\_info(['get\_current\_time','convert\_time'])" \} \} </tool\_call>}\\

Calling a tool after fetching the schema: \\
\texttt{<tool\_call> \{ "name": "code\_interpreter", "arguments": \{ "code": "time\_mcp.get\_current\_time(timezone='America/New\_York')" \} \} </tool\_call>}\\

\end{tcolorbox}

\end{document}